\documentclass{article}

\usepackage{microtype}

\usepackage{subfig}
\usepackage{caption}
\usepackage{listings}

\usepackage{booktabs} 

\usepackage{amsmath,amsfonts,bm}

\def\eqref#1{equation~\ref{#1}}

\def\1{\bm{1}}

\def\rvx{{\mathbf{x}}}

\def\rvz{{\mathbf{z}}}

\def\va{{\bm{a}}}

\def\vc{{\bm{c}}}

\def\vo{{\bm{o}}}
\def\vp{{\bm{p}}}
\def\vq{{\bm{q}}}

\def\vs{{\bm{s}}}

\def\vu{{\bm{u}}}

\def\vw{{\bm{w}}}
\def\vx{{\bm{x}}}
\def\vy{{\bm{y}}}
\def\vz{{\bm{z}}}

\def\mB{{\bm{B}}}

\def\mQ{{\bm{Q}}}

\def\mX{{\bm{X}}}

\def\mZ{{\bm{Z}}}

\DeclareMathAlphabet{\mathsfit}{\encodingdefault}{\sfdefault}{m}{sl}
\SetMathAlphabet{\mathsfit}{bold}{\encodingdefault}{\sfdefault}{bx}{n}

\def\gA{{\mathcal{A}}}

\def\gD{{\mathcal{D}}}

\def\gH{{\mathcal{H}}}

\def\gL{{\mathcal{L}}}
\def\gM{{\mathcal{M}}}
\def\gN{{\mathcal{N}}}
\def\gO{{\mathcal{O}}}

\def\gS{{\mathcal{S}}}

\def\gX{{\mathcal{X}}}

\newcommand{\E}{\mathbb{E}}

\newcommand{\R}{\mathbb{R}}

\newcommand{\KL}{D_{\mathrm{KL}}}

\usepackage{graphicx}
\graphicspath{{./figs/}}

\usepackage{hyperref}
\usepackage[capitalise,nameinlink,noabbrev]{cleveref}

\newcommand{\xxnote}[3]{}
\ifx\hidenotes\undefined
  \renewcommand{\xxnote}[3]{\color{#2}{#1: #3}}
\fi

\newcommand{\proto}{\textbf{Proto-RL}}
\newcommand{\protos}{Proto-RL}

\usepackage[accepted]{icml2021}

\icmltitlerunning{Reinforcement Learning with Prototypical Representations}

\begin{document}

\twocolumn[
\icmltitle{Reinforcement Learning with Prototypical Representations}




\begin{icmlauthorlist}
\icmlauthor{Denis Yarats}{nyu,fair}
\icmlauthor{Rob Fergus}{nyu}
\icmlauthor{Alessandro Lazaric}{fair}
\icmlauthor{Lerrel Pinto}{nyu}
\end{icmlauthorlist}

\icmlaffiliation{nyu}{New York University}
\icmlaffiliation{fair}{Facebook AI Research}

\icmlcorrespondingauthor{Denis Yarats}{denisyarats@cs.nyu.edu}

\icmlkeywords{Machine Learning, ICML}

\vskip 0.3in
]



\printAffiliationsAndNotice{} 

\newif\ifincludeappendix

\begin{abstract}
Learning effective representations in image-based environments is crucial for sample efficient Reinforcement Learning (RL). Unfortunately, in RL, representation learning is confounded with the exploratory experience of the agent -- learning a useful representation requires diverse data, while effective exploration is only possible with coherent representations. Furthermore, we would like to learn representations that not only generalize across tasks but also accelerate downstream exploration for efficient task-specific training. To address these challenges we propose \proto, a self-supervised framework that ties representation learning with exploration through prototypical representations. These prototypes simultaneously serve as a summarization of the exploratory experience of an agent as well as a basis for representing observations. We pre-train these task-agnostic representations and prototypes on environments without downstream task information. This enables state-of-the-art downstream policy learning on a set of difficult continuous control tasks. We open-source our code at~\url{https://github.com/denisyarats/proto}.
\end{abstract}

\section{Introduction}

Reinforcement Learning (RL) with rich visual observations has proven to be a recipe for success in a variety of domains ranging from gameplay~\cite{mnih2013dqn,silver2016go} to robotics~\cite{levine2015e2etraining}. A crucial ingredient for successful image-based RL is to learn an encoder that maps the high-dimensional input to a compact representation capturing the \textit{latent} state of the environment. Standard RL methods can then be applied using the latent representation to efficiently learn policies. Unfortunately, representation learning in RL poses several challenges.

First, fitting encoders using the scarce supervisory signal from rewards alone is sample inefficient and leads to poor performance. Prior work \cite{srinivas2020curl,yarats2019improving} addresses this problem by leveraging self-supervised techniques alongside standard image-based RL, which leads to more robust and effective representations. However, such techniques are not effective in settings where task-specific reward, which drives exploration, is absent. 


\begin{figure*}[t!]
    \centering
    
    \subfloat[Evolution of the state-visitation distribution and prototypes during the task-agnostic pre-training phase of \protos.]{\includegraphics[width=0.9\linewidth]{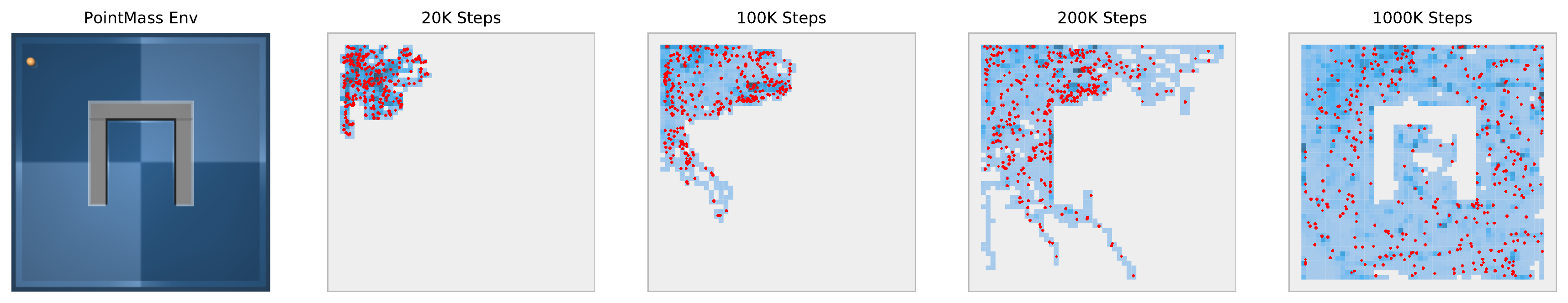}  \label{fig:pm_expl}}\\
 
    \subfloat[Evolution of the state-visitation distribution during the downstream RL phase of \protos~for the Reach Center task.]{\includegraphics[width=0.9\linewidth]{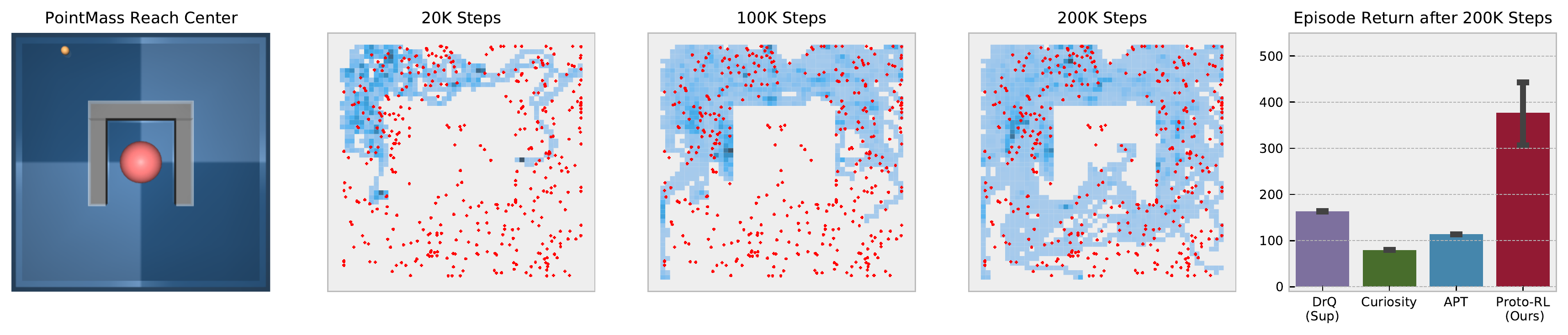}  \label{fig:pm_tunel}}\\

         \caption{An example of~\proto~running on the pixel-based U-maze pointmass environment with {\bf(a)} task-agnostic pre-training, followed by {\bf (b)} downstream RL. {\bf (a)}: task-agnostic exploration and representation learning stage. The state-visitation distribution is shown in blue, which converges to uniform coverage with sufficient steps. Red points depict the prototypes via closest states in the embedding space. {\bf (b)}: subsequent application to the Reach Center task with sparse reward. Note the rapid exploration of the environment facilitated by the pre-trained prototypes (and embedding function). \protos~discovers the goal location in only 200k steps, while other methods struggle to solve the task. The experiment details are provided in~\cref{section:point_mass}.}

    \label{fig:pm}
\end{figure*}

A second, more fundamental challenge is that in the context of RL, representation learning is intimately connected to the exploration of the environment and vice-versa~\cite{sekar2020planning,liu2020apt}. The data observed by the agent is non-stationary and depends on the regions of state space covered during exploration. If exploration is ineffective, the latent space produced by the encoder cannot properly characterize all parts of the environment, degrading performance for downstream tasks. Conversely, the exploration strategy cannot be defined directly on input images since no algorithm would be able to exhaustively explore all possible images. Hence, a representation accurately capturing the latent state of the environment is needed so that the agent can distinguish novel latent states from those already visited and focus exploration on the former. This leads to a chicken and egg problem, where learning useful representations requires diverse data, while effective exploration is only possible with coherent representations.

Finally, a desirable property of latent representations is to generalize across tasks defined in the same environment. This requires the representation to support a wide range of policies. Furthermore, the representation should also facilitate exploration for new tasks through the organization inherent in the latent space. 
Recent approaches to representation learning in RL \cite{srinivas2020curl,laskin2020reinforcement,yarats2021image} are effective for a specific task but the representations obtained are intrinsically tied to it and often
require to re-learn representations under a different objective in cases where the desired behavior induces a previously unseen state visitation distribution.

In this work, we address the three challenges described above through \proto, a framework for image-based RL that learns self-supervised visual representations without access to task-specific rewards. Concretely, we consider the few-shot unsupervised RL setting, which consists of two distinct phases. In the first phase, the agent explores an environment in a task-agnostic fashion to learn its visual representations. Then in the second phase, given the learned representations, the agent is required to solve a downstream task with as few environment interactions as possible. Such a setup evaluates the agent's ability to quickly solve a new task without any prior knowledge of it.

During the first phase \protos~learns an encoder to embed visual observations in a low-dimensional latent space, along with a set of prototypical embeddings, which we refer to as \textit{prototypes}~\citep{asano2020selflabelling,caron2021unsupervised} that form the basis of this latent space. 
To effectively explore the environment in the absence of task rewards, \protos~trains a policy that maximizes intrinsic reward measured by particle-based entropy~\citep{singh203knnent}. The latent embeddings and prototypes are trained together on observations from the exploration policy. The policy receives intrinsic reward computed using current prototypes to encourage the visitation of unexplored regions of state space. During the second phase \protos~uses the pre-trained encoder along with the prototypes to accelerate RL for downstream tasks. 

\cref{fig:pm} illustrates the behavior of \protos~in an image-based navigation task. In particular, it shows the effectiveness of the unsupervised exploration strategy to thoroughly cover the state space, thus providing a diverse enough dataset for representation learning. \protos~also returns discrete prototypes that are evenly spread over the state space and are used to improve exploration during the downstream stage. An accurate representation of the latent state, together with efficient prototype-based exploration, leads \protos~to achieve state-of-the-art performance. 


To summarize, this paper makes the following contributions: (i) We propose a novel task-agnostic pre-training scheme that learns an embedding function, along with a set of prototypical representations directly from visual observations; (ii) we demonstrate the ability of the learned representations and prototypes to generalize to unseen downstream tasks from the DeepMind Control Suite~\citep{tassa2018dmcontrol}, with significant improvements over current state-of-the-art methods; (iii) we show that the prototypes and learned representation enables efficient downstream exploration, especially in sparse reward settings.

\section{Related Work}
In this section we provide a brief description on the most relevant work and ideas that \protos~builds on top of.

\paragraph{Self-supervised learning in Computer Vision~(CV)}
Self-supervision has proven to be an effective technique to learn representations from large amounts of unlabeled data~\cite{vincent2008extracting, doersch2015unsupervised,Wang_UnsupICCV2015, noroozi2016unsupervised,zhang2017split,gidaris2018unsupervised}. Several creative ideas have been used to self-supervise such as video tracking~\cite{Wang2019Learning}, augmentation prediction~\cite{chen2020simple}, and puzzle solving~\cite{noroozi2016unsupervised} among others. Such pre-trained representations provide a strong initialization for downstream finetuning for tasks such as image classification~\cite{chen2020simple, hnaff2019dataefficient, wu2018unsupervised, he2019momentum}. \protos~is partly inspired by SwAV~\citep{caron2021unsupervised}, where prototypical representations are learned through contrastive losses~\cite{oord2018representation} that encourage consistency across random augmentations of the input. However, unlike SwAV that learns on a stationary dataset of images, \protos~operates in dynamic environments in an RL setting that inherently produces non-stationary data distributions for learning. 

\paragraph{Representation learning in RL}
To enable sample efficient RL from pixels, several researchers have taken inspiration from the successes of representation learning in computer vision and looked at learning coherent latent representations alongside RL.  SAC-AE~\citep{yarats2019improving}, SLAC~\cite{lee2019stochastic}, demonstrated how auto-encoders can be used to learn representations that improve RL. Following this,  CURL~\citep{srinivas2020curl}, SPR~\cite{schwarzer2020data}, ATC~\cite{stooke2020decoupling} used losses that encourage consistency across random observational augmentations to further improve sample-efficiency. More recently, data augmentations by themselves have shown significant success in learning representations~\cite{laskin2020reinforcement,kostrikov2020image}. Model-based RL has also looked at learning these representations from predictive losses~\cite{hafner2018planet,hafner2019dream,yan2020learning,finn2015deepspatialae,pinto2016curious,agrawal2016learning}. Finally, several works use auxiliary losses derived from state or demonstrations to learn representations~\cite{jaderberg2016reinforcement,zhan2020framework,young2020visual,chen2020robust}. We note that these works in general focus on learning continuous representations of the environment through interactive experience without explicitly encouraging exploration. In contrast, \protos~not only learns representations on interactive experience, but also uses prototypes for better exploration. 

\paragraph{Exploration and Intrinsic Motivation in RL}
A fundamental problem in RL is exploring the state space of the underlying MDP, especially in cases where the reward is sparse or absent. Approaches that tackle this problem are generally task-agnostic and exploit various inductive biases that correlate positively with efficient exploration. Prior approaches include using state visitation counts~\citep{bellemare2016unifying,ostrovski2017countbased}, curiosity-driven exploration~\citep{pathak2017curiositydriven}, distilling random networks~\cite{burda2018exploration}, hindsight relabeling~\citep{andrychowicz2017hindsight}, state visitation entropy maximization~\citep{hazan2019provably, mutti2020policy,liu2020apt}, ensemble disagreement~\citep{sekar2020planning}, among others.  \protos~builds on these ideas and focuses on exploration by maximizing the entropy of the state visitation distribution~\citep{hazan2019provably}. 
However, in contrast to prior work~\citep{liu2020apt}, \protos~uses prototypical representations to better estimate entropy, which improves downstream exploration.

\section{Background}

\subsection{Task-Agnostic RL from Images}
\label{section:task_agnostic}
We formulate task-agnostic image-based control as an infinite-horizon partially observable Markov Decision Process (POMDP)~\citep{bellman1957mdp,kaelbling1998planning} without rewards, as a tuple $\gM = (\mathbf{\gO}, \gA, P, \gamma, d_0)$, where $\gO$ is the high-dimensional observation space (image pixels), $\gA$ is the action space, $P: \gO^* \times \gA \to \Delta(\gO)$ is the transition function\footnote{We denote by $\gO^*$ an arbitrarily long sequence of observations and by $\Delta(\gO)$ a distribution over the space of observations $\gO$.} that defines a probability distribution over the next observation given the sequence of past observations and the current action, $\gamma \in [0, 1)$ is a discount factor, and $d_0 \in \Delta(\gO)$ is the distribution of the initial observation $\vo_0$. 
Per common practice~\citep{mnih2013dqn}, throughout the paper the task-agnostic POMDP is converted into a task-agnostic MDP~\citep{bellman1957mdp} $(\gX, \gA, P, \gamma, d_0)$ by stacking three consecutive previous image observations into a trajectory snippet $\vx_t = \{\vo_t, \vo_{t-1}, \vo_{t-2}\}$ and defining the corresponding state space $\gX$ and the transition function $\vx_{t+1} \sim P(\cdot|\vx_t,\va_{t})$. Any policy $\pi : \gX \to \Delta(\gA)$ induces discounted state visitation distribution $d^\pi(\vx) =  (1-\gamma) \sum_{t=0}^\infty \gamma^t d_t^\pi(\vx)$, where $d_t^\pi(\vx) = P(\vx_t = \vx| \vx_0 \sim d_0, \forall t'<t, \va_{t'} \sim \pi(\cdot|\vx_{t'}), \vx_{t'+1} \sim P(\cdot|\vx_{t'}, \va_{t'}))$. Similar to \citet{hazan2019provably,lee2019efficient,mutti2020policy}, we focus on the exploratory goal of finding the policy $\pi$ that maximizes the entropy $\mathbb{H}(d^\pi) = -\sum_{\vx} d^\pi(\vx) \log (d^\pi(\vx))$ of the state visitation distribution.

\subsection{Task-Specific RL from Images}
\label{section:task_specific}
In the downstream RL setup the reward-free MDP is extended with a reward function $R: \gS \times \gA \to \mathbb{R}$ to form the task-specific MDP $(\gX, \gA, P, R, \gamma, d_0)$. The objective then is to find a policy $\pi : \gX \to \Delta(\gA)$ to maximize the expected discounted sum of rewards $\E_\pi[\sum_{t=0}^\infty \gamma^t r_t]$, where  $\vx_0 \sim d_0$, and $\forall t$ we have $\va_{t} \sim \pi(\cdot|\vx_{t})$, $\vx_{t+1} \sim P(\cdot| \vx_{t}, \va_{t})$, and $r_t = R(\vx_{t}, \va_{t})$.





\begin{figure*}[t!]
    \centering

    \includegraphics[width=\linewidth]{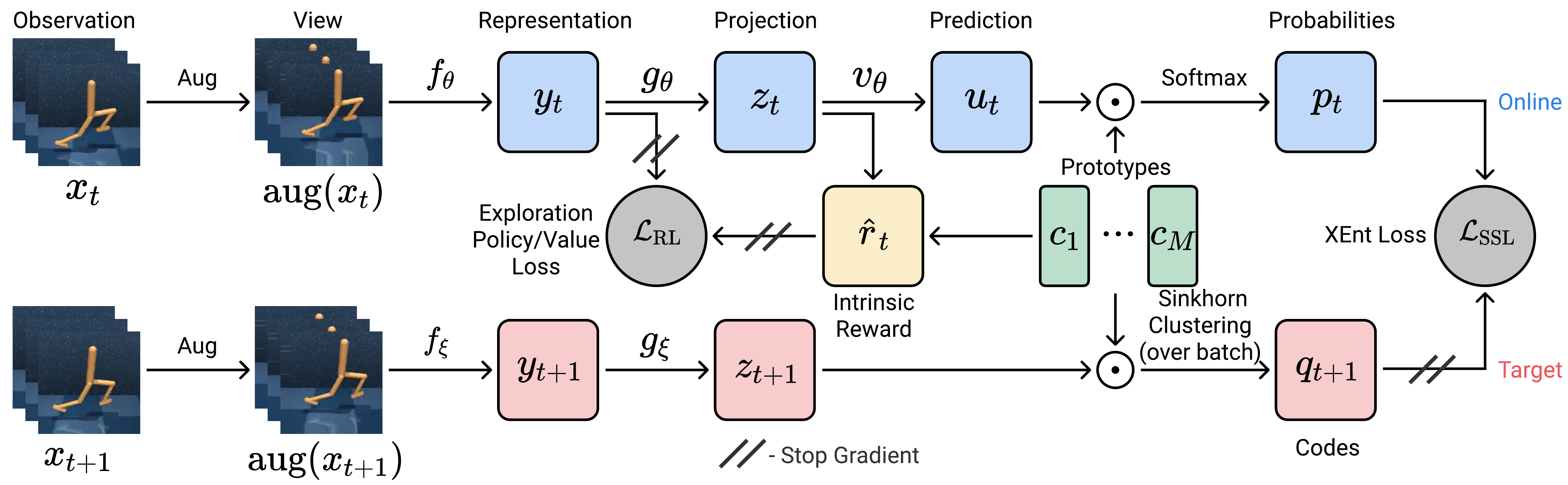}
    \caption{\protos~proposes a self-supervised scheme that learns to encode high-dimensional image observations $\vx_t$, $\vx_{t+1}$, using an encoder $f_\theta$ along with a set of prototypes $\{{\vc_i}\}_{i=1}^M$ that defines the basis of the latent space. Learning is done by optimizing the clustering assignment loss $\mathcal{L}_{\mathrm{SSL}}$. To encourage exploration, prototypes are simultaneously used to compute an entropy-based intrinsic reward $\hat{r}_t$ that is maximized by the exploration agent. To decouple representation learning from the exploration task, we block the gradients of the agent's RL loss $\mathcal{L}_{\mathrm{RL}}$ from updating the encoder and prototypes. See~\cref{section:method} for a full description.}
    \label{fig:proto_model}
\end{figure*}

\subsection{Nearest Neighbor Entropy Estimation}
Estimation of entropy for a distribution $p(\gX)$ defined on a $q$-dimensional space $\gX \subseteq \mathbb{R}^q$ is often done via Monte Carlo using a finite set of samples $\mX =\{\vx_i\}_{i=1}^N \sim p(\gX)$ to obtain $\hat{\mathbb{H}}_\mX(p)= -\frac{1}{N}\sum_{i=1}^N \log p(\vx_i)$. However, this estimator requires the ability to not only sample from $p$, but also to estimate pointwise density. This is often intractable in high-dimensional continuous spaces, such as those in image-based RL. An alternative approach is to use a non-parametric Nearest Neighbor (NN) based entropy estimator~\cite{singh203knnent}:
\begin{align*}
    \hat{\mathbb{H}}_{k,\mX}(p) &= -\frac{1}{N}\sum_{i=1}^N\ln\frac{k\Gamma(q/2+1)}{N \pi^{q/2} R_{i,k,\mX}^q } + C_k,
\end{align*}
where $\Gamma$ is the gamma function, $C_k=\ln k -\frac{\Gamma'(k)}{\Gamma(k)}$ is the bias correction term, and  $R_{i,k,\mX}=\|\vx_i - \mathrm{NN}_{k,\mX}(\vx_i)\|$ is the Euclidean distance between $\vx_i$ and its $k^{\text{th}}$ nearest neighbor from the dataset $\mX$, defined as $\mathrm{NN}_{k,\mX}(\vx_i)$.

Going forward, we are only interested in the proportional estimation of entropy that simplifies the estimator:
\begin{align}
\label{eqn:entropy}
    \hat{\mathbb{H}}_{k,\mX}(p) &\propto \sum_{i=1}^N \ln \|
    \vx_i - \mathrm{NN}_{k,\mX}(\vx_i)\|.
\end{align}
Here, each point $\vx_i$ contributes an amount proportional to $\|
    \vx_i - \mathrm{NN}_{k,\mX}(\vx_i)\|$ to the total entropy of the dataset $\mX$.
This estimator is shown to be asymptotically unbiased and consistent~\citep{singh203knnent}.

\section{Proto-RL Algorithm}
\label{section:method}

In this section, we provide technical details on \protos. Our goal is to learn visual representations in a task-agnostic fashion through interactions with a task-agnostic POMDP~(see~\cref{section:task_agnostic} for terminology and setup). To do this, we design a self-supervised scheme that fits an encoder that embeds high-dimensional observations to low-dimensional latent states and defines an exploration strategy that allows for the discovery of diverse transitions. In~\cref{section:proto_rep_learning}, we describe our representation learning framework, which focuses on learning prototypes that form the basis for our visual embeddings. In~\cref{section:proto_exploration}, we describe how these prototypes yield a metric that enables entropy-based exploration. Both the representations and the exploration are learned simultaneously and are collectively referred to as \protos, which is summarized in~
\cref{section:proto_pretraining}. Once the representations are learned, we describe how they can accelerate downstream learning of tasks in~\cref{section:proto_downstream}. 




\subsection{Prototypical Representation Learning}
\label{section:proto_rep_learning}

Our framework learns a visual encoder that maps pixels to continuous latent embeddings, as well as a basis within this latent space, as defined by a set of prototypical vectors. Our novel self-supervised scheme trains the encoder and prototypes simultaneously by projecting observation encodings onto clusters (prototypes) and comparing them with cluster assignment targets. These are produced by a projection of the encodings of the next observation onto the prototypes, constrained to ensure uniform prototype coverage over the dataset. Our approach draws inspiration from the recent CV approach SwAV~\citep{caron2021unsupervised}, adapting these ideas to the non-stationary RL setting.



The \protos~framework, illustrated in~\cref{fig:proto_model}, computes representations as follows. The augmented input frames $\vx_t$ are mapped to a continuous embedding $\vy_t$ using the convolutional image encoder $f_\theta$. $\vy_t$ then undergoes a projection by the MLP network $g_\theta$ to produce latent vector $\vz_t$, and then another MLP network $v_\theta$ to attain features $\vu_t$. The final step is to project $\vu_t$ into a basis defined by a set of $M$ continuous vectors $\{\vc_i\}_{i=1}^M$, which we call prototypes. This is done by using a softmax to produce $\vp_t$, a probability vector over the $M$ prototypes whose components are:
\begin{align*}
    p_t^{(i)} &= \frac{\exp (\hat{\vu}_t^T \vc_i / \tau)}{\sum_{k=1}^M \exp(\hat{\vu}_t^T \vc_k / \tau)}, \text{ where } \hat{\vu}_t = \frac{\vu_t}{\|\vu_t\|_2}
\end{align*}
and $\tau$ is the softmax temperature hyper-parameter.

Learning involves simultaneously training (1) the encoder $f_\theta$, (2) the projector $g_\theta$, (3) the predictor $v_\theta$ and (4) the prototype vectors $\{\vc_i\}_{i=1}^M$. These form the \emph{online} network. To optimize the \emph{online} parameters, a \emph{target} network is used to  produce a target probability vector $\vq_{t+1}$. 
The \emph{target} network inputs the next augmented observation $\vx_{t+1}$ and encodes it using the target encoder $f_\xi$ to produce continuous embedding $\vy_{t+1}$, then $\vy_{t+1}$ is fed to the target projector $g_\xi$ to produce latent encodings $\vz_{t+1}$. These target projections are then used to compute a target probability vector $\vq_{t+1}$ using the  prototypes. Intuitively, the vector $\vq_{t+1}$ represents the soft assignment of the target embedding to the prototypes. To ensure equal partitioning of the prototypes across all embeddings, we employ the Sinkhorn-Knopp clustering procedure~\citep{cuturi2013sinkhorn,caron2021unsupervised}, which is run over a mini-batch of embeddings.  This clustering procedure constrains each prototype to be assigned to the same number of samples in the batch while maintaining complete coverage.
Operationally, given a batch size of $B$, the Sinkhorn-Knopp procedure begins with a $M \times B$ matrix with each element initialized to $\hat{\vz}^T_{t+1,b}\vc_m$, where $\hat{\vz}=\vz/\|\vz\|_2$. It then iteratively produces a doubly-normalized matrix, the columns of which comprise $\vq_{t+1}$ for the batch. 
The corresponding $\vp_t$ and $\vq_{t+1}$ are then used to compute a cross-entropy loss:
\begin{align*}
    \gL_{\mathrm{SSL}}(\vp_t, \vq_{t+1}) &= - \vq_{t+1}^T \log \vp_{t}. 
\end{align*}
Importantly, the gradient of this loss is only used to update the online network parameters (e.g. $\theta$ and $\{\vc_i\}_{i=1}^M$), while being blocked in the target network. The weights of the target network $\xi$ are instead updated using the exponential moving average of online network weights $\theta$. Note that this update, and the use of predictor $v_\theta$ introduces an asymmetry between the two networks that prevents collapse to trivial solutions~\cite{grill2020bootstrap}. The pseudo-code for our framework is provided in~\cref{section:pseudocode}. 

\paragraph{Architectures} The online and target encoders $f_\theta$ and $f_\xi$ both use the architecture from SAC-AE~\citep{yarats2019improving}. The online and target projectors $g_\theta$ and $g_\xi$ are linear layers with $128$ outputs. The online predictor $v_\theta$ is a 2-layer MLP with ReLU non-linearities.~\protos~learns $M=512$  prototypes, each parameterized as a $128$-dimensional vector.
\paragraph{Data} The learning framework described above implements a novel contrastive scheme which compares views of two consecutive observations $\vx_t$ and $\vx_{t+1}$, augmented with random image shifts~\cite{yarats2021image}. This differs from other representation learning for RL approaches such as CURL~\citep{srinivas2020curl}, which contrasts two different views of the same observation $\vx_t$, and ATC~\citep{stooke2020decoupling}, which uses temporal contrast over a trajectory snippet. 
As mentioned in \cref{section:task_agnostic}, the input $\vx$ consists of a stack of three image frames. New data is gathered via an unsupervised exploration policy that uses current embeddings $\vy_t$, projections $\vz_t$ and prototypes $\{\vc_i\}_{i=1}^M$, which we detail in the next section.

\begin{figure}[t!]
    \centering

    \includegraphics[width=\linewidth]{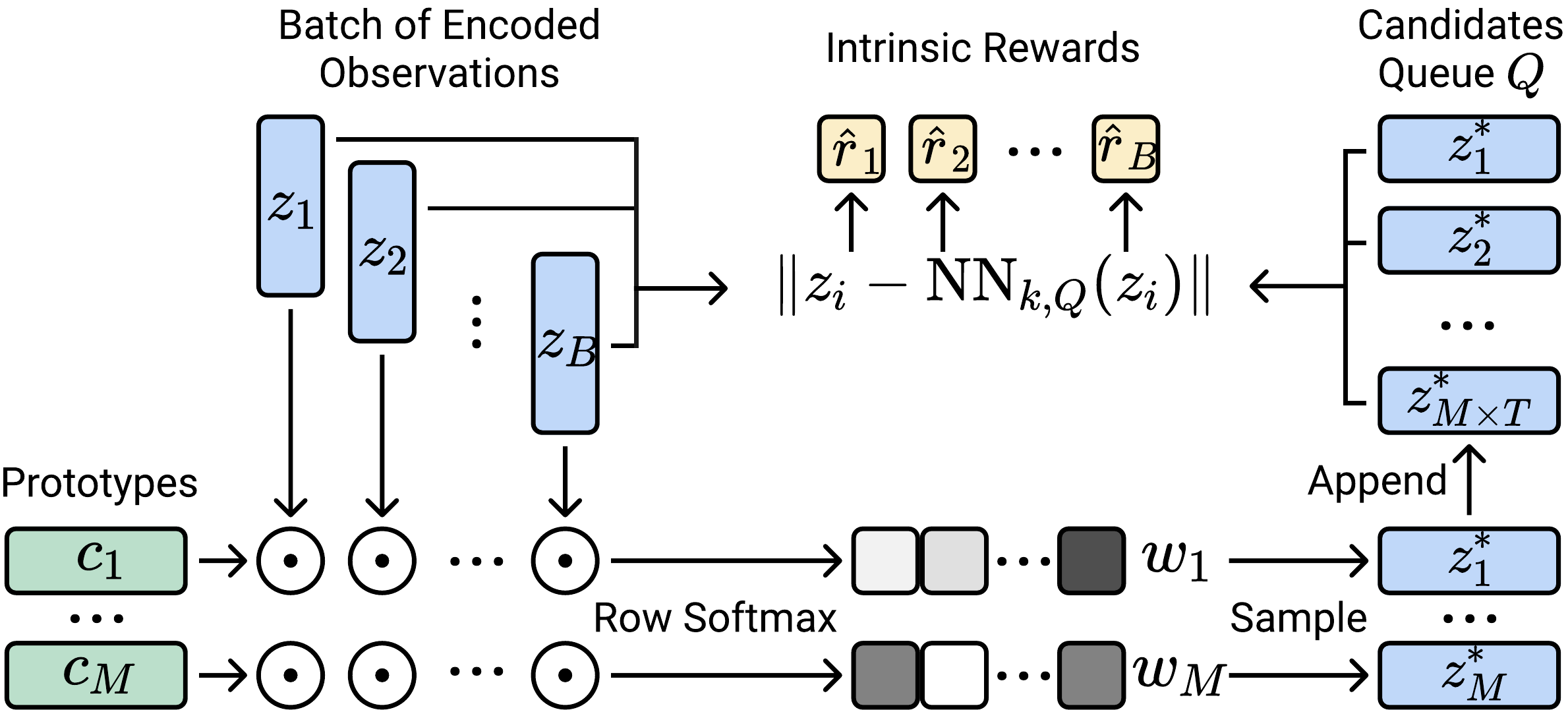}
    \caption{The entropy-based intrinsic reward used by~\protos. This employs a nearest-neighbor estimator (\cref{eqn:entropy})  computed over a set of embeddings $\mQ$ that are uniformly drawn from clustering of a batch of encoded observations $\{\vz_i\}_{i=1}^B$ with the current prototypes $\{{\vc_i}\}_{i=1}^M$. See~\cref{section:proto_exploration} for more details.}
    \label{fig:proto_reward}
\end{figure}

\subsection{Maximum Entropy Exploration}
\label{section:proto_exploration}


\begin{figure*}[t!]
    \centering

    \includegraphics[width=\linewidth]{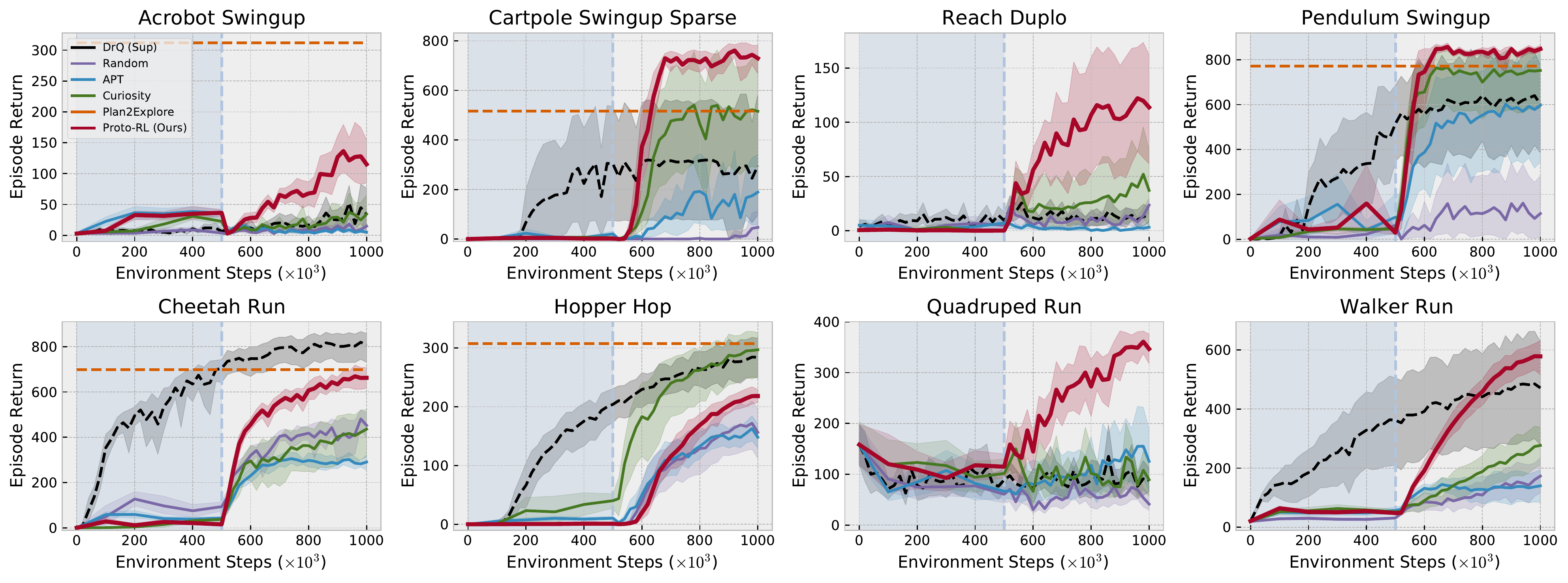}
    \caption{Single task evaluation using eight challenging environments from DeepMind Control Suite. For each method (except for DrQ and Plan2Explore), we first perform task-agnostic pretraining for 500k environment steps, before introducing task reward and training for a further 500k steps. DrQ uses task reward from the outset. Plan2Explore\footnotemark[2], being model-based, uses an intermediate methodology, described in~\cref{subsection:exp_setup}. \protos~consistently beats the baselines and in many cases exceeds the fully supervised approach of DrQ.  }
    \label{fig:single_task}
\end{figure*}


When reward signal is absent and no assumptions about the MDP can be made, one possible intrinsic objective for the agent is to learn a policy which maximizes the entropy $\hat{\mathbb{H}}_{k,\mX\sim d^{\pi}(\cdot)}(d^\pi)$ of the discounted state visitation distribution $d^\pi(\rvx)$, per ~\cref{eqn:entropy}.
Although the estimator is  asymptotically unbiased and consistent~\citep{singh203knnent}, applying it in practice poses several challenges that we address using the learned encoder and prototypes (\cref{section:proto_rep_learning}). 

First, estimation in the original high-dimensional image space $\gX$ is a poor metric for measuring similarity. To this end, we estimate entropy using Euclidean distance to the  $k^{\text{th}}$ nearest neighbor in the low-dimensional learned latent space:  $\hat{\mathbb{H}}_{k,\mZ\sim d^{\pi}(\cdot)}(d^\pi)  \propto \sum_{i=1}^N \ln \|
    \vz_i - \mathrm{NN}_{k,\mZ}(\vz_i)\|$, where $\vz_i = g_\theta(f_\theta (\vx_i))$ and $\mZ=\{\vz_i\}_{i=1}^N$.

Second, finding the  $k^{\text{th}}$ nearest neighbor over the entire dataset $\mZ$ becomes computationally expensive as the dataset grows in size. One possible solution, proposed by~\citet{liu2020apt}, is to constrain the search to a random batch $\mB$ of embeddings  uniformly drawn from the replay buffer $\mZ$ as $\hat{\mathbb{H}}_{k,\mB \sim \mZ}(d^{\pi})$. Empirically, this approximation leads to a high variance estimate. For example, in a recently discovered part of the environment the state density in the buffer will be low, thus the estimated distance will be unduly large. To mitigate the problem of under-representation of novel states, we use a dataset rebalancing scheme that up-weights novel embeddings and down-weights common ones. This is done by clustering the candidates using the learned prototypes and then uniformly sampling from these clusters. To implement this for each prototype $\vc_j$ we compute a softmax distribution $\vw_j$ with components $w_j^{(i)} = \frac{\exp (\hat{\vz_i}^T \vc_j)}{ \sum_{k=1}^B \exp (\hat{\vz_k}^T \vc_j)}$  over a batch of $L_2$ normalized projections  $\{\hat{\vz}_i\}_{i=1}^B$ and then sample a constant number of candidates from this distribution. The sampled candidates are stored in a queue $\mQ$ of a fixed size $M \times T$, where $T$ candidates are used per cluster. 

Finally,~\protos~ modifies an original task-agnostic transition $(\vx_t, \va_t, \vx_{t+1})$ by encoding visual observations with the online encoder $f_\theta$ into embeddings $\vy_t = f_\theta (\vx_t)$ and $\vy_{t+1}=f_\theta (\vx_{t+1})$, then adding the entropy-based intrinsic reward computed using the candidate set $\mQ$ as:
\begin{align}
    \hat{r}_t &= \| \vz_{t+1} - \mathrm{NN}_{k,\mQ}(\vz_{t+1})\|.\label{eqn:max_ent_reward}
\end{align} 
The proposed scheme is visualized in~\cref{fig:proto_reward}.

\subsection{Pretraining with Task-Agnostic RL}
\label{section:proto_pretraining}
To collect a diverse dataset for enabling representation learning,~\protos~simultaneously trains an exploration RL agent to optimize the intrinsic reward specified by~\cref{eqn:max_ent_reward}. The RL agent is trained on transitions $(\vy_t, \va_t, \hat{r}_t, \vy_{t+1})$ described above. Importantly, We block the gradients from the RL loss $\mathcal{L}_{\mathrm{RL}}$, defined in~\cref{section:extended_background}, in order to learn task-agnostic representations and prototypes. The RL agent is implemented using SAC~\citep{haarnoja2018sac}.


\subsection{Application to Downstream Tasks}
\label{section:proto_downstream}
To perform downstream RL training we (i) use the online encoder $f_\theta$ to map image observations $\vx_t, \vx_{t+1}$ into embeddings $\vy_t, \vy_{t+1}$ and (ii) augment the extrinsic reward $r_t$ with the intrinsic reward $\hat{r}_t$, scaled by hyper-parameter $\alpha$. This results into the modified transitions $(\vy_t, \va_t, r_t + \alpha \hat{r}_t, \vy_{t+1})$, which are then used to train a standard state-based RL algorithm. In this work, since we are interested in studying the effects of the task-agnostic representation alone, we freeze the encoder and prototypes. However, we note that finetuning representations and prototypes during downstream RL is also compatible with our framework.


\section{Experiments}
In this section we discuss empirical results on using \protos{}~for learning visual representations. We begin by describing our experimental setup and evaluation protocols. We then use this setting to answer the following questions: (a) Does task-agnostic pre-training improve downstream task-specific RL? (b) How well do the learned representations transfer to different tasks? (c) How important is exploration during representation learning? (d) Can the pre-trained prototypes be used to improve downstream exploration? (e) Does \protos{} explore the state space more effectively than baselines? (f) Do the learned prototypes provide more accurate entropy estimation?


\subsection{Experimental Setup}
\label{subsection:exp_setup}
Our agents operate in the few-shot unsupervised RL setting with two learning phases. In the task-agnostic phase the agent is allowed to interact with an environment, 
but it does not have access to any information about the downstream task that the agent will be asked to solve in the next phase. In the downstream RL phase, rewards associated with a task are revealed to the agent. 
In our experiments, agents are allowed 500k environment interactions in the task-agnostic phase, followed by 500k additional interactions with the environment in the downstream RL phase.

\paragraph{Environment Details} We use the DeepMind Control Suite~\citep{tassa2018dmcontrol}, a challenging benchmark for image-based RL. Following prior work, visual observations are represented as $84 \times 84 \times 3$ pixel renderings. The episode length is $1000$ for all tasks, except \textit{Reach Duplo}, where it is $250$. A fixed action repeat $R=2$~\citep{hafner2019dream} is applied across all environments. Each agent's performance is evaluated over $10$ episodes every $10000$ environment steps. All figures plot the mean performance over $10$ random seeds, together with $\pm 1$ standard deviation shading. 
\paragraph{Hyper-parameters} \protos~is trained using Adam~\citep{kingma2014adam} with learning rate $10^{-4}$ and mini-batch size of $512$. The downstream exploration hyper-parameter is $\alpha=0.2$ and the number of cluster candidates is set to $T=4$. We use SAC implementation from~\citet{yarats2020pytorch_sac}.  

\begin{figure}[t!]
    \centering

    \includegraphics[width=\linewidth]{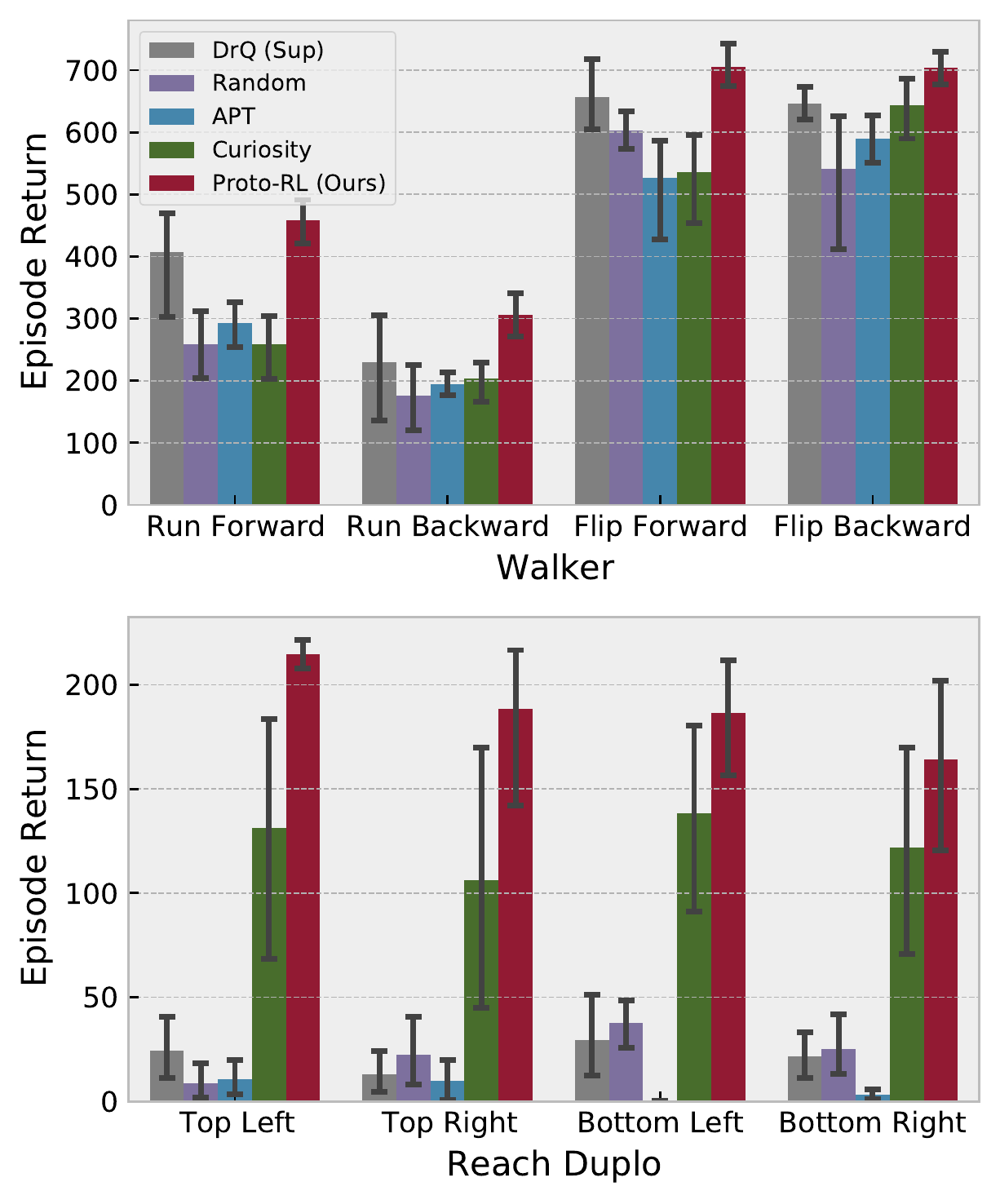}
    \vspace{-5mm}
    \caption{Multi-task evaluation using two domains from DeepMind Control Suite, with four tasks in each. We perform task-agnostic pre-training for 500k steps in each domain. The frozen representation and prototypes are then applied separately to each of the four tasks, training for additional 500k steps with the task reward. DrQ performance is measured after training for 500k steps. The results show that the representations learned by \protos~generalize well and enable efficient learning of multiple downstream tasks.  
    }
    \label{fig:multitask}
\end{figure}

\begin{figure*}[t!]
    \centering

    \includegraphics[width=\linewidth]{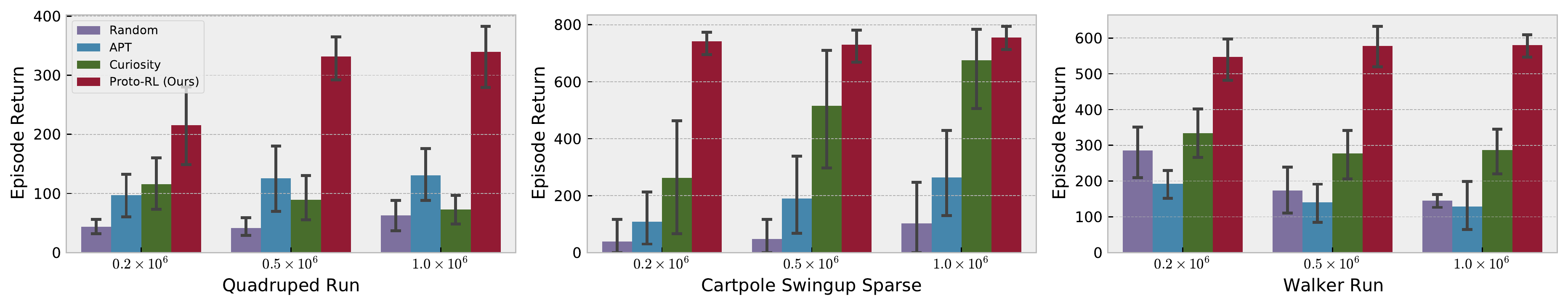}
    \caption{Varying the amount of task-agnostic pre-training on three different tasks from DeepMind Control Suite. Subsequent task-specific training (on top of the frozen representation) uses 500k steps. \protos~is able to explore state space sufficiently within 200k steps to learn representations that can support downstream tasks. }
    \label{fig:different_expl}
\end{figure*}

\paragraph{Baselines}
To contextualize the results of \protos, we compare with the following baseline algorithms:
\begin{itemize}
    \item Random exploration: The agent is based on DrQ~\citep{yarats2021image} and it explores the environment using a random policy during the task-agnostic phase. We then freeze the learned encoder to provide representations for task-specific RL training.
    \item Curiosity~\cite{pathakICMl17curiosity}: This agent explores the environment using a curiosity-driven intrinsic motivation reward along with learning continuous visual representations using DrQ.
    \item APT~\cite{liu2020apt}: The agent explores the environment using an entropy-driven intrinsic motivation reward along with learning continuous visual representations. Since APT does not use  prototypical representations, entropy is measured through sampling of observations from the replay buffer.
    \item Plan2Explore~\cite{sekar2020planning}: Here, a model-based algorithm Dreamer~\citep{hafner2019dream} is used in conjunction with Curiosity~\cite{pathakICMl17curiosity} to explore the environment. However, since Plan2Explore uses model-based optimization while \protos~and other baselines are model-free, we only denote the final performance of Plan2Explore to avoid ambiguities in step-wise comparisons. Furthermore, Plan2Explore is provided with an estimate of the reward function of the downstream task at the end of the task-agnostic pre-training, which allows it to leverage the model to plan directly for a task-specific policy in a zero-shot manner. On the other hand, all other baselines have to learn the reward function directly in downstream RL. 
    \item DrQ~\cite{yarats2021image}: Here, a state-of-the-art method for task-specific RL is trained on task-specific rewards for 1M steps to anchor the performance ranges.
\end{itemize}

The full experimental setup and details on baselines are described in~\cref{section:hyperparams}.

\subsection{Task-Agnostic Pre-training}
\label{subsection:task_agnostic}
We present results on eight environments in~\cref{fig:single_task}, with extended results on sixteen environments in~\cref{section:full_pretrainig}. \protos~significantly improves upon Random exploration and APT across all environments, while being better than Curiosity based exploration in 7/8 environments. This demonstrates that in the context of model-free RL, \protos~provides state-of-the-art downstream task learning. Furthermore, \protos~trained on 500k task-agnostic environment interactions achieves competitive performance to the model-based algorithm Plan2Explore\footnote{\label{p2e_note}Note that we only compare on environments reported in the original paper since the publicly released code underperforms the reported numbers.} that is trained on 1M unsupervised steps, followed by the 200k fine-tuning steps with reward.

Perhaps a more exciting result is that \protos~trained with 500k steps of downstream RL outperforms DrQ trained on 1M steps in 6/8 environments. This demonstrates how task-agnostic representation learning can enable superior downstream RL and achieve state-of-the-art image-based RL results. Note, that these environments are indeed among the hardest image-based environments from DeepMind Control Suite~\citep{tassa2018dmcontrol}. In~\cref{section:no_pretraining}, we also show that while~\protos~can outperform baselines when no prior task-agnostic pre-training is done, our task-agnostic pre-training scheme provides additional improvement.


\subsection{Multi-Task Generalization}
As pointed out in the introduction, one desirable property of task-agnostic representations is that they can effectively generalize across different downstream tasks defined in the same environment.
To highlight this ability of \protos, we present results on RL training on different downstream tasks 
in~\cref{fig:multitask}. After 500k steps of downstream RL,~\protos~significantly outperforms all the baselines we compare with. We believe this ability of prototypes to accelerate downstream task learning through both better representations and exploration is key to unlock more effective and robust generalization in image-based RL tasks, as it is case in computer vision. Details about the multi-task environments are provided in~\cref{section:multi_task}.


\begin{figure}[t!]
    \centering

    \includegraphics[width=\linewidth]{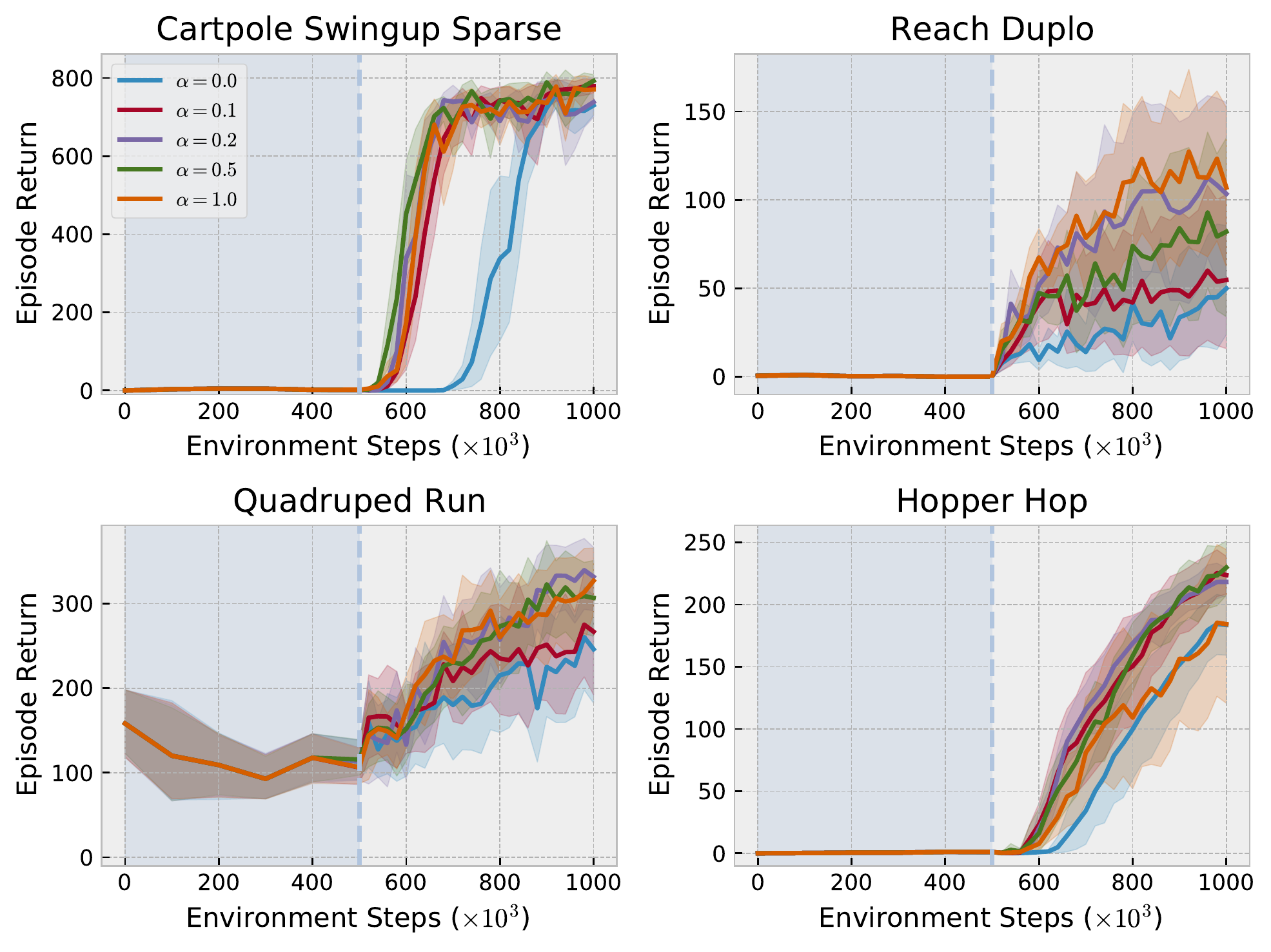}
    \caption{Evaluation regime from~\cref{fig:single_task} for \protos~but varying the balance between exploration and downstream reward using hyper-parameter $\alpha$. We see that $\alpha > 0$ facilitates learning, especially in the sparse reward tasks such as \textit{Cartpole Swingup Sparse} and \textit{Reach Duplo}.}
    \label{fig:downstream_expl}
\end{figure}

\subsection{Efficiency of Task-Agnostic Pre-training}
\label{subsection:eff_of_pretraining}
In the previous experiments we describe the performance of various unsupervised RL algorithms on the 500k task-agnostic steps benchmark. However, this raises a question on how many such unsupervised steps are required to learn representations that can accelerate downstream tasks. In~\cref{fig:different_expl} we present comparisons on 200k, 500k, and 1M steps of task-agnostic training. We find that \protos~consistently outperforms the baselines across the various splits. Interestingly, on \textit{Walker Run} we see that for the baselines, performance drops with increased task-agnostic training, which highlights the difficulty in learning generalizable representations without overfitting to the explored data. 

\subsection{Downstream Exploration}
\label{subsection:downstream_exploration}
A key differentiating factor of \protos~compared to current relevant methods is that prototypes enable exploration even during downstream task RL. To understand the importance of this, we study of effect of the hyper-parameter $\alpha$ that trades off the task reward with our entropy-based intrinsic reward in~\cref{fig:downstream_expl}. Across all the tasks, using $\alpha=0$ i.e., not using the prototype-driven exploration, underperforms every experiment that uses $\alpha \geq 0.1$. Notably, for sparse reward tasks like \textit{Cartpole Swingup Sparse} and \textit{Reach Duplo} $\alpha \geq 0.1$ is significantly better that $\alpha=0$. This highlights the importance of using  prototypes that summarize the exploratory experience in an environment. Interestingly, even without using the prototype-driven exploration, \protos~is able to solve all tasks with a performance that is still competitive with the baselines. This shows that the image embeddings learned during the task-agnostic pre-training are indeed effective in solving the downstream tasks.


\subsection{State Coverage of Task-Agnostic Pre-training}
\begin{figure}[t!]
    \centering

    \includegraphics[width=\linewidth]{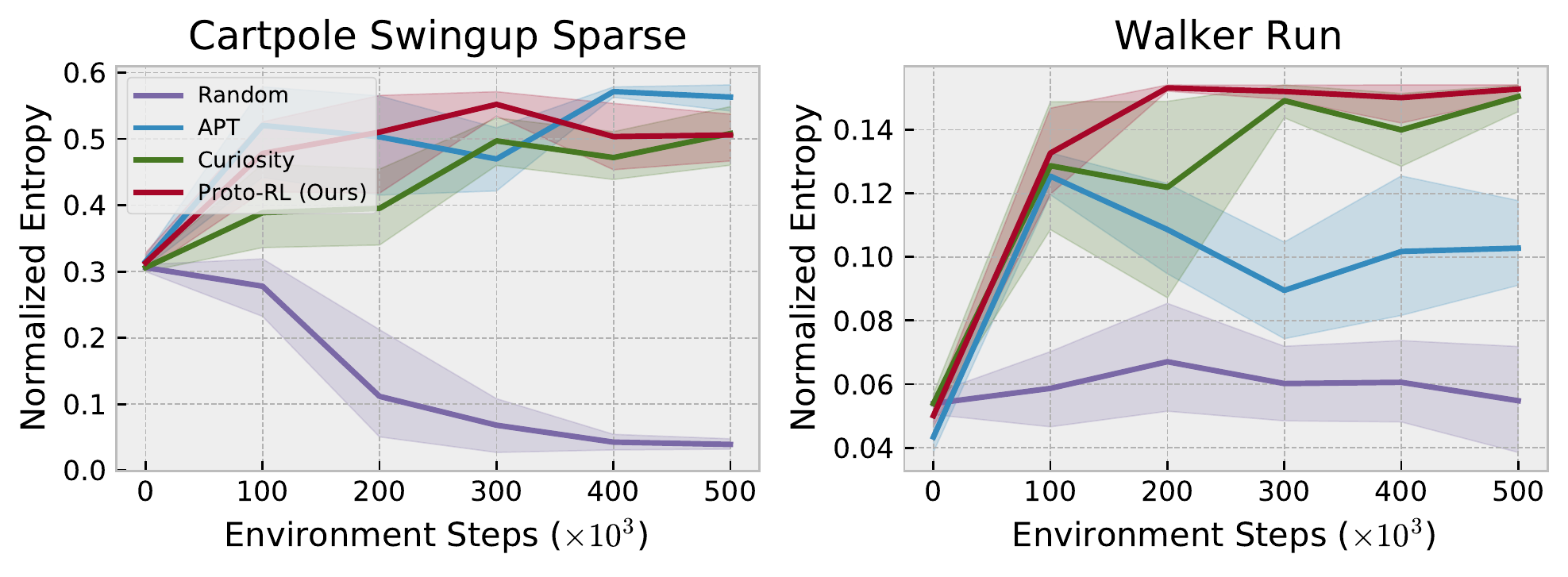}
    \caption{Proto-RL demonstrates wider state coverage during the task-agnostic pre-training stage compared to the other baselines by achieving a state visitation distribution of higher entropy.}
    \label{fig:expl_entropy}
\end{figure}

In this section we evaluate Proto-RL's ability to maximize entropy of the state visitation distribution induced by an exploration policy during the task-agnostic pre-training phase. Following~\citet{hazan2019provably}, we compute the normalized entropy by discretizing the continuous state space and continuously averaging these discrete entropies for each state stored in the replay buffer as training progresses. Finally, these entropy estimates are normalized by the maximum possible entropy and plotted in~\cref{fig:expl_entropy}. 

\subsection{Variance of State Entropy Estimates}
\begin{figure}[t!]
    \centering

    \includegraphics[width=\linewidth]{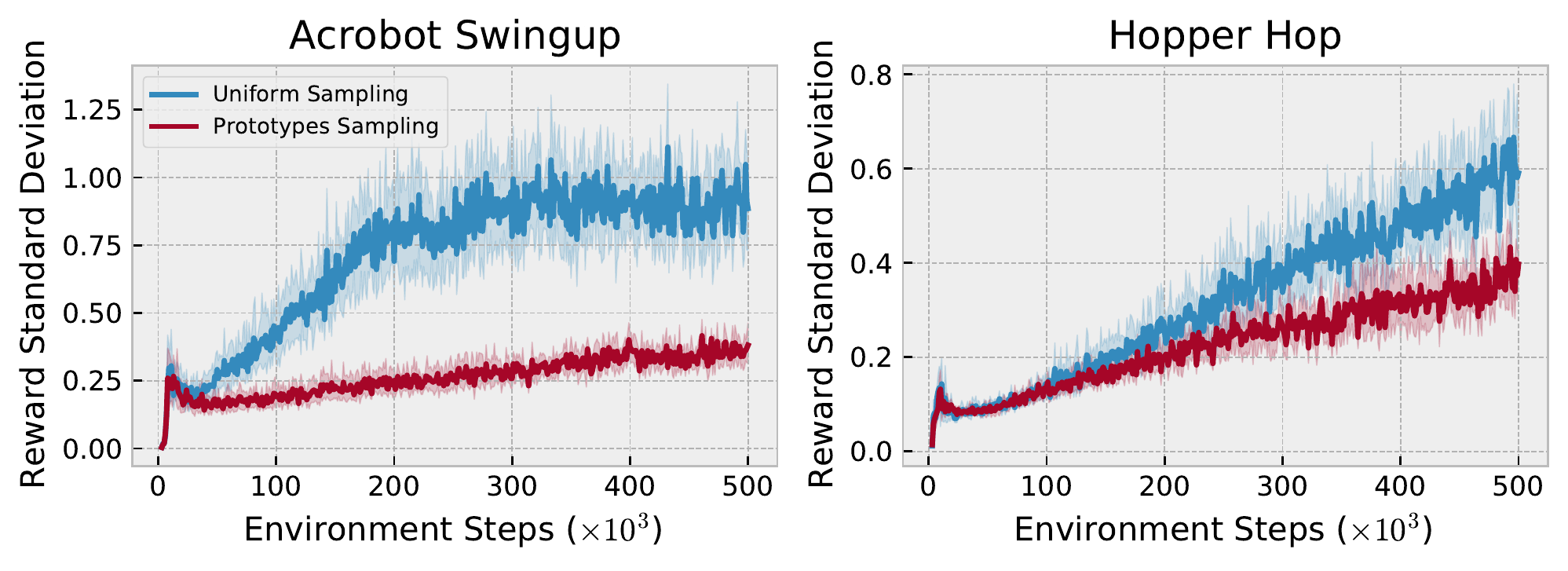}
    \caption{Our prototype-based candidates sampling scheme demonstrates more stable nearest neighbor estimates of entropy comparing to uniform sampling as done in~\citet{liu2020apt} .}
    \label{fig:reward_var}
\end{figure}

\protos~relies on an accurate estimation of entropy of the state visitation distribution which is computed by sampling candidates $Q_{\mathrm{proto}}$ from each of the clusters implicitly defined by the prototypes. Compared to computing nearest neighbors using a random set of candidates $Q_{\mathrm{uniform}}$, our approach demonstrates lower variance. To gain some intuition here, let's consider the case where we are evaluating the intrinsic reward for a state $\rvz$ in a mid-to-low density region, depending on the composition of the random candidate set, we may obtain either a very large reward (i.e., if $Q_{\mathrm{uniform}}$ fails at containing any of the samples close to $\rvz$) or very low (i.e., if $Q_{\mathrm{uniform}}$ contains mostly samples close to $\rvz$). On the other hand, by using uniformly spread prototypes, we always have a few candidates in $Q_{\mathrm{proto}}$ that are close to $\rvz$ as well as far from it, which leads to a more stable entropy estimation. This in turn impacts the stability of learning from the intrinsic rewards and eventually leads to a better exploration strategy. This effect is demonstrated in~\cref{fig:reward_var}.

\section{Conclusion}
In this paper we present \protos, an unsupervised representation learning algorithm for RL. \protos~simultaneously learns representations and prototypes from visual inputs while exploring environments in a task-agnostic fashion. Empirically, the learned representations and prototypes enable state-of-the-art exploration and learning of downstream objectives, as well as effective generalization across multiple tasks. We believe \protos~brings us a step closer to ``fine-tuning" in RL, a process that is commonplace in modern computer vision and natural language processing. It also opens up several directions for future research such as understanding the theoretical underpinnings of discrete representations, applications to robotics and offline RL. 

\section{Acknowledgements}
This research is supported in part by DARPA through the
Machine Common Sense Program. We thank Oleh Rybkin and Danijar Hafner for sharing data for baselines. We also thank Hao Liu and anonymous reviewers for insightful feedback that helped to improve our paper.

\newpage 

\bibliography{main}

\begin{thebibliography}{60}
\providecommand{\natexlab}[1]{#1}
\providecommand{\url}[1]{\texttt{#1}}
\expandafter\ifx\csname urlstyle\endcsname\relax
  \providecommand{\doi}[1]{doi: #1}\else
  \providecommand{\doi}{doi: \begingroup \urlstyle{rm}\Url}\fi

\bibitem[Agrawal et~al.(2016)Agrawal, Nair, Abbeel, Malik, and
  Levine]{agrawal2016learning}
Agrawal, P., Nair, A., Abbeel, P., Malik, J., and Levine, S.
\newblock Learning to poke by poking: experiential learning of intuitive
  physics.
\newblock In \emph{Proceedings of the 30th International Conference on Neural
  Information Processing Systems}, pp.\  5092--5100, 2016.

\bibitem[Andrychowicz et~al.(2017)Andrychowicz, Wolski, Ray, Schneider, Fong,
  Welinder, McGrew, Tobin, Abbeel, and Zaremba]{andrychowicz2017hindsight}
Andrychowicz, M., Wolski, F., Ray, A., Schneider, J., Fong, R., Welinder, P.,
  McGrew, B., Tobin, J., Abbeel, O.~P., and Zaremba, W.
\newblock Hindsight experience replay.
\newblock In \emph{Advances in neural information processing systems}, pp.\
  5048--5058, 2017.

\bibitem[Asano et~al.(2020)Asano, Rupprecht, and
  Vedaldi]{asano2020selflabelling}
Asano, Y.~M., Rupprecht, C., and Vedaldi, A.
\newblock Self-labelling via simultaneous clustering and representation
  learning.
\newblock \emph{arXiv preprint arXiv}, 2020.

\bibitem[Ba et~al.(2016)Ba, Kiros, and Hinton]{ba2016layernorm}
Ba, J.~L., Kiros, J.~R., and Hinton, G.~E.
\newblock Layer normalization.
\newblock \emph{arXiv e-prints}, 2016.

\bibitem[Bellemare et~al.(2016)Bellemare, Srinivasan, Ostrovski, Schaul,
  Saxton, and Munos]{bellemare2016unifying}
Bellemare, M.~G., Srinivasan, S., Ostrovski, G., Schaul, T., Saxton, D., and
  Munos, R.
\newblock Unifying count-based exploration and intrinsic motivation.
\newblock \emph{arXiv preprint arXiv}, 2016.

\bibitem[Bellman(1957)]{bellman1957mdp}
Bellman, R.
\newblock A markovian decision process.
\newblock \emph{Indiana Univ. Math. J.}, 1957.

\bibitem[Burda et~al.(2018)Burda, Edwards, Storkey, and
  Klimov]{burda2018exploration}
Burda, Y., Edwards, H., Storkey, A., and Klimov, O.
\newblock Exploration by random network distillation.
\newblock \emph{arXiv preprint arXiv}, 2018.

\bibitem[Caron et~al.(2021)Caron, Misra, Mairal, Goyal, Bojanowski, and
  Joulin]{caron2021unsupervised}
Caron, M., Misra, I., Mairal, J., Goyal, P., Bojanowski, P., and Joulin, A.
\newblock Unsupervised learning of visual features by contrasting cluster
  assignments, 2021.

\bibitem[Chen et~al.(2020{\natexlab{a}})Chen, Sax, Lewis, Armeni, Savarese,
  Zamir, Malik, and Pinto]{chen2020robust}
Chen, B., Sax, A., Lewis, G., Armeni, I., Savarese, S., Zamir, A., Malik, J.,
  and Pinto, L.
\newblock Robust policies via mid-level visual representations: An experimental
  study in manipulation and navigation.
\newblock \emph{arXiv preprint arXiv:2011.06698}, 2020{\natexlab{a}}.

\bibitem[Chen et~al.(2020{\natexlab{b}})Chen, Kornblith, Norouzi, and
  Hinton]{chen2020simple}
Chen, T., Kornblith, S., Norouzi, M., and Hinton, G.
\newblock A simple framework for contrastive learning of visual
  representations.
\newblock \emph{arXiv preprint arXiv:2002.05709}, 2020{\natexlab{b}}.

\bibitem[Cuturi(2013)]{cuturi2013sinkhorn}
Cuturi, M.
\newblock Sinkhorn distances: Lightspeed computation of optimal transport.
\newblock In \emph{NIPS}, 2013.

\bibitem[Doersch et~al.(2015)Doersch, Gupta, and
  Efros]{doersch2015unsupervised}
Doersch, C., Gupta, A., and Efros, A.~A.
\newblock Unsupervised visual representation learning by context prediction.
\newblock In \emph{Proceedings of the IEEE International Conference on Computer
  Vision}, pp.\  1422--1430, 2015.

\bibitem[Finn et~al.(2015)Finn, Tan, Duan, Darrell, Levine, and
  Abbeel]{finn2015deepspatialae}
Finn, C., Tan, X.~Y., Duan, Y., Darrell, T., Levine, S., and Abbeel, P.
\newblock Learning visual feature spaces for robotic manipulation with deep
  spatial autoencoders.
\newblock \emph{CoRR}, 2015.

\bibitem[Fujimoto et~al.(2018)Fujimoto, van Hoof, and Meger]{fujimoto2018td3}
Fujimoto, S., van Hoof, H., and Meger, D.
\newblock Addressing function approximation error in actor-critic methods.
\newblock In \emph{Proceedings of the 35th International Conference on Machine
  Learning, {ICML} 2018, Stockholmsmassan, Stockholm, Sweden, July 10-15,
  2018}, 2018.

\bibitem[Gidaris et~al.(2018)Gidaris, Singh, and
  Komodakis]{gidaris2018unsupervised}
Gidaris, S., Singh, P., and Komodakis, N.
\newblock Unsupervised representation learning by predicting image rotations,
  2018.

\bibitem[Grill et~al.(2020)Grill, Strub, Altché, Tallec, Richemond,
  Buchatskaya, Doersch, Pires, Guo, Azar, Piot, Kavukcuoglu, Munos, and
  Valko]{grill2020bootstrap}
Grill, J.-B., Strub, F., Altché, F., Tallec, C., Richemond, P.~H.,
  Buchatskaya, E., Doersch, C., Pires, B.~A., Guo, Z.~D., Azar, M.~G., Piot,
  B., Kavukcuoglu, K., Munos, R., and Valko, M.
\newblock Bootstrap your own latent: A new approach to self-supervised
  learning, 2020.

\bibitem[Haarnoja et~al.(2018)Haarnoja, Zhou, Hartikainen, Tucker, Ha, Tan,
  Kumar, Zhu, Gupta, Abbeel, et~al.]{haarnoja2018sac}
Haarnoja, T., Zhou, A., Hartikainen, K., Tucker, G., Ha, S., Tan, J., Kumar,
  V., Zhu, H., Gupta, A., Abbeel, P., et~al.
\newblock Soft actor-critic algorithms and applications.
\newblock \emph{arXiv preprint arXiv:1812.05905}, 2018.

\bibitem[Hafner et~al.(2018)Hafner, Lillicrap, Fischer, Villegas, Ha, Lee, and
  Davidson]{hafner2018planet}
Hafner, D., Lillicrap, T., Fischer, I., Villegas, R., Ha, D., Lee, H., and
  Davidson, J.
\newblock Learning latent dynamics for planning from pixels.
\newblock \emph{arXiv preprint arXiv:1811.04551}, 2018.

\bibitem[Hafner et~al.(2019)Hafner, Lillicrap, Ba, and
  Norouzi]{hafner2019dream}
Hafner, D., Lillicrap, T., Ba, J., and Norouzi, M.
\newblock Dream to control: Learning behaviors by latent imagination.
\newblock \emph{arXiv preprint arXiv:1912.01603}, 2019.

\bibitem[Hazan et~al.(2019)Hazan, Kakade, Singh, and Soest]{hazan2019provably}
Hazan, E., Kakade, S.~M., Singh, K., and Soest, A.~V.
\newblock Provably efficient maximum entropy exploration, 2019.

\bibitem[He et~al.(2020)He, Fan, Wu, Xie, and Girshick]{he2019momentum}
He, K., Fan, H., Wu, Y., Xie, S., and Girshick, R.
\newblock Momentum contrast for unsupervised visual representation learning.
\newblock In \emph{Proceedings of the IEEE Conference on Computer Vision and
  Pattern Recognition}, 2020.

\bibitem[Hénaff et~al.(2019)Hénaff, Srinivas, Fauw, Razavi, Doersch, Eslami,
  and van~den Oord]{hnaff2019dataefficient}
Hénaff, O.~J., Srinivas, A., Fauw, J.~D., Razavi, A., Doersch, C., Eslami, S.
  M.~A., and van~den Oord, A.
\newblock Data-efficient image recognition with contrastive predictive coding,
  2019.

\bibitem[Jaderberg et~al.(2016)Jaderberg, Mnih, Czarnecki, Schaul, Leibo,
  Silver, and Kavukcuoglu]{jaderberg2016reinforcement}
Jaderberg, M., Mnih, V., Czarnecki, W.~M., Schaul, T., Leibo, J.~Z., Silver,
  D., and Kavukcuoglu, K.
\newblock Reinforcement learning with unsupervised auxiliary tasks, 2016.

\bibitem[Kaelbling et~al.(1998)Kaelbling, Littman, and
  Cassandra]{kaelbling1998planning}
Kaelbling, L.~P., Littman, M.~L., and Cassandra, A.~R.
\newblock Planning and acting in partially observable stochastic domains.
\newblock \emph{Artificial intelligence}, 1998.

\bibitem[Kingma \& Ba(2014)Kingma and Ba]{kingma2014adam}
Kingma, D.~P. and Ba, J.
\newblock Adam: A method for stochastic optimization.
\newblock \emph{arXiv preprint arXiv:1412.6980}, 2014.

\bibitem[Kostrikov et~al.(2020)Kostrikov, Yarats, and
  Fergus]{kostrikov2020image}
Kostrikov, I., Yarats, D., and Fergus, R.
\newblock Image augmentation is all you need: Regularizing deep reinforcement
  learning from pixels.
\newblock 2020.

\bibitem[Laskin et~al.(2020)Laskin, Lee, Stooke, Pinto, Abbeel, and
  Srinivas]{laskin2020reinforcement}
Laskin, M., Lee, K., Stooke, A., Pinto, L., Abbeel, P., and Srinivas, A.
\newblock Reinforcement learning with augmented data, 2020.

\bibitem[Lee et~al.(2019{\natexlab{a}})Lee, Nagabandi, Abbeel, and
  Levine]{lee2019stochastic}
Lee, A.~X., Nagabandi, A., Abbeel, P., and Levine, S.
\newblock Stochastic latent actor-critic: Deep reinforcement learning with a
  latent variable model.
\newblock \emph{arXiv preprint arXiv:1907.00953}, 2019{\natexlab{a}}.

\bibitem[Lee et~al.(2019{\natexlab{b}})Lee, Eysenbach, Parisotto, Xing, Levine,
  and Salakhutdinov]{lee2019efficient}
Lee, L., Eysenbach, B., Parisotto, E., Xing, E.~P., Levine, S., and
  Salakhutdinov, R.
\newblock Efficient exploration via state marginal matching.
\newblock \emph{CoRR}, abs/1906.05274, 2019{\natexlab{b}}.
\newblock URL \url{http://arxiv.org/abs/1906.05274}.

\bibitem[Levine et~al.(2015)Levine, Finn, Darrell, and
  Abbeel]{levine2015e2etraining}
Levine, S., Finn, C., Darrell, T., and Abbeel, P.
\newblock End-to-end training of deep visuomotor policies.
\newblock \emph{CoRR}, abs/1504.00702, 2015.

\bibitem[Lillicrap et~al.(2015)Lillicrap, Hunt, Pritzel, Heess, Erez, Tassa,
  Silver, and Wierstra]{lillicrap2015ddpg}
Lillicrap, T.~P., Hunt, J.~J., Pritzel, A., Heess, N., Erez, T., Tassa, Y.,
  Silver, D., and Wierstra, D.
\newblock Continuous control with deep reinforcement learning.
\newblock \emph{CoRR}, 2015.

\bibitem[Liu \& Abbeel(2021)Liu and Abbeel]{liu2020apt}
Liu, H. and Abbeel, P.
\newblock Unsupervised active pre-training for reinforcement learning.
\newblock \emph{openreview}, 2021.
\newblock URL \url{https://openreview.net/forum?id=cvNYovr16SB}.

\bibitem[Mnih et~al.(2013)Mnih, Kavukcuoglu, Silver, Graves, Antonoglou,
  Wierstra, and Riedmiller]{mnih2013dqn}
Mnih, V., Kavukcuoglu, K., Silver, D., Graves, A., Antonoglou, I., Wierstra,
  D., and Riedmiller, M.
\newblock Playing atari with deep reinforcement learning.
\newblock \emph{arXiv e-prints}, 2013.

\bibitem[Mutti et~al.(2021)Mutti, Pratissoli, and Restelli]{mutti2020policy}
Mutti, M., Pratissoli, L., and Restelli, M.
\newblock Task-agnostic exploration via policy gradient of a non-parametric
  state entropy estimate.
\newblock In \emph{Proceedings of the AAAI Conference on Artificial
  Intelligence}, volume~35, pp.\  9028--9036, 2021.

\bibitem[Noroozi \& Favaro(2016)Noroozi and Favaro]{noroozi2016unsupervised}
Noroozi, M. and Favaro, P.
\newblock Unsupervised learning of visual representations by solving jigsaw
  puzzles.
\newblock In \emph{European Conference on Computer Vision}, pp.\  69--84.
  Springer, 2016.

\bibitem[Ostrovski et~al.(2017)Ostrovski, Bellemare, van~den Oord, and
  Munos]{ostrovski2017countbased}
Ostrovski, G., Bellemare, M.~G., van~den Oord, A., and Munos, R.
\newblock Count-based exploration with neural density models.
\newblock \emph{arXiv preprint arXiv}, 2017.

\bibitem[Pathak et~al.(2017{\natexlab{a}})Pathak, Agrawal, Efros, and
  Darrell]{pathak2017curiositydriven}
Pathak, D., Agrawal, P., Efros, A.~A., and Darrell, T.
\newblock Curiosity-driven exploration by self-supervised prediction.
\newblock \emph{arXiv preprint arXiv}, 2017{\natexlab{a}}.

\bibitem[Pathak et~al.(2017{\natexlab{b}})Pathak, Agrawal, Efros, and
  Darrell]{pathakICMl17curiosity}
Pathak, D., Agrawal, P., Efros, A.~A., and Darrell, T.
\newblock Curiosity-driven exploration by self-supervised prediction.
\newblock In \emph{ICML}, 2017{\natexlab{b}}.

\bibitem[Pinto et~al.(2016)Pinto, Gandhi, Han, Park, and
  Gupta]{pinto2016curious}
Pinto, L., Gandhi, D., Han, Y., Park, Y.-L., and Gupta, A.
\newblock The curious robot: Learning visual representations via physical
  interactions.
\newblock In \emph{European Conference on Computer Vision}, pp.\  3--18.
  Springer, 2016.

\bibitem[Schwarzer et~al.(2020)Schwarzer, Anand, Goel, Hjelm, Courville, and
  Bachman]{schwarzer2020data}
Schwarzer, M., Anand, A., Goel, R., Hjelm, R.~D., Courville, A., and Bachman,
  P.
\newblock Data-efficient reinforcement learning with momentum predictive
  representations.
\newblock \emph{arXiv preprint arXiv:2007.05929}, 2020.

\bibitem[Sekar et~al.(2020)Sekar, Rybkin, Daniilidis, Abbeel, Hafner, and
  Pathak]{sekar2020planning}
Sekar, R., Rybkin, O., Daniilidis, K., Abbeel, P., Hafner, D., and Pathak, D.
\newblock Planning to explore via self-supervised world models.
\newblock \emph{arXiv preprint arXiv}, 2020.

\bibitem[Silver et~al.(2016)Silver, Huang, Maddison, Guez, Sifre, van~den
  Driessche, Schrittwieser, Antonoglou, Panneershelvam, Lanctot, Dieleman,
  Grewe, Nham, Kalchbrenner, Sutskever, Lillicrap, Leach, Kavukcuoglu, Graepel,
  and Hassabis]{silver2016go}
Silver, D., Huang, A., Maddison, C.~J., Guez, A., Sifre, L., van~den Driessche,
  G., Schrittwieser, J., Antonoglou, I., Panneershelvam, V., Lanctot, M.,
  Dieleman, S., Grewe, D., Nham, J., Kalchbrenner, N., Sutskever, I.,
  Lillicrap, T., Leach, M., Kavukcuoglu, K., Graepel, T., and Hassabis, D.
\newblock Mastering the game of go with deep neural networks and tree search.
\newblock \emph{Nature}, 529:\penalty0 484--503, 2016.
\newblock URL
  \url{http://www.nature.com/nature/journal/v529/n7587/full/nature16961.html}.

\bibitem[Singh et~al.(2003)Singh, Misra, Hnizdo, Fedorowicz, and
  Demchuk]{singh203knnent}
Singh, H., Misra, N., Hnizdo, V., Fedorowicz, A., and Demchuk, E.
\newblock Nearest neighbor estimates of entropy.
\newblock \emph{American Journal of Mathematical and Management Sciences},
  23\penalty0 (3-4):\penalty0 301--321, 2003.

\bibitem[Srinivas et~al.(2020)Srinivas, Laskin, and Abbeel]{srinivas2020curl}
Srinivas, A., Laskin, M., and Abbeel, P.
\newblock Curl: Contrastive unsupervised representations for reinforcement
  learning.
\newblock \emph{arXiv preprint arXiv:2004.04136}, 2020.

\bibitem[Stooke et~al.(2020)Stooke, Lee, Abbeel, and
  Laskin]{stooke2020decoupling}
Stooke, A., Lee, K., Abbeel, P., and Laskin, M.
\newblock Decoupling representation learning from reinforcement learning.
\newblock \emph{arXiv preprint arXiv}, 2020.

\bibitem[Tassa et~al.(2018)Tassa, Doron, Muldal, Erez, Li, Casas, Budden,
  Abdolmaleki, Merel, Lefrancq, et~al.]{tassa2018dmcontrol}
Tassa, Y., Doron, Y., Muldal, A., Erez, T., Li, Y., Casas, D. d.~L., Budden,
  D., Abdolmaleki, A., Merel, J., Lefrancq, A., et~al.
\newblock Deepmind control suite.
\newblock \emph{arXiv preprint arXiv:1801.00690}, 2018.

\bibitem[van~den Oord et~al.(2018)van~den Oord, Li, and
  Vinyals]{oord2018representation}
van~den Oord, A., Li, Y., and Vinyals, O.
\newblock Representation learning with contrastive predictive coding, 2018.

\bibitem[van Hasselt et~al.(2015)van Hasselt, Guez, and
  Silver]{hasselt2015doubledqn}
van Hasselt, H., Guez, A., and Silver, D.
\newblock Deep reinforcement learning with double q-learning.
\newblock \emph{arXiv e-prints}, 2015.

\bibitem[Vincent et~al.(2008)Vincent, Larochelle, Bengio, and
  Manzagol]{vincent2008extracting}
Vincent, P., Larochelle, H., Bengio, Y., and Manzagol, P.-A.
\newblock Extracting and composing robust features with denoising autoencoders.
\newblock In \emph{Proceedings of the 25th international conference on Machine
  learning}, pp.\  1096--1103. ACM, 2008.

\bibitem[Wang \& Gupta(2015)Wang and Gupta]{Wang_UnsupICCV2015}
Wang, X. and Gupta, A.
\newblock Unsupervised learning of visual representations using videos.
\newblock In \emph{ICCV}, 2015.

\bibitem[Wang et~al.(2019)Wang, Jabri, and Efros]{Wang2019Learning}
Wang, X., Jabri, A., and Efros, A.~A.
\newblock Learning correspondence from the cycle-consistency of time.
\newblock In \emph{CVPR}, 2019.

\bibitem[Wu et~al.(2018)Wu, Xiong, Yu, and Lin]{wu2018unsupervised}
Wu, Z., Xiong, Y., Yu, S.~X., and Lin, D.
\newblock Unsupervised feature learning via non-parametric instance
  discrimination.
\newblock In \emph{Proceedings of the IEEE Conference on Computer Vision and
  Pattern Recognition}, pp.\  3733--3742, 2018.

\bibitem[Yan et~al.(2020)Yan, Vangipuram, Abbeel, and Pinto]{yan2020learning}
Yan, W., Vangipuram, A., Abbeel, P., and Pinto, L.
\newblock Learning predictive representations for deformable objects using
  contrastive estimation.
\newblock \emph{arXiv preprint arXiv:2003.05436}, 2020.

\bibitem[Yarats \& Kostrikov(2020)Yarats and Kostrikov]{yarats2020pytorch_sac}
Yarats, D. and Kostrikov, I.
\newblock Soft actor-critic (sac) implementation in pytorch.
\newblock \url{https://github.com/denisyarats/pytorch_sac}, 2020.

\bibitem[Yarats et~al.(2019)Yarats, Zhang, Kostrikov, Amos, Pineau, and
  Fergus]{yarats2019improving}
Yarats, D., Zhang, A., Kostrikov, I., Amos, B., Pineau, J., and Fergus, R.
\newblock Improving sample efficiency in model-free reinforcement learning from
  images.
\newblock \emph{arXiv preprint arXiv:1910.01741}, 2019.

\bibitem[Yarats et~al.(2021)Yarats, Kostrikov, and Fergus]{yarats2021image}
Yarats, D., Kostrikov, I., and Fergus, R.
\newblock Image augmentation is all you need: Regularizing deep reinforcement
  learning from pixels.
\newblock In \emph{9th International Conference on Learning Representations,
  ICLR 2021}, 2021.

\bibitem[Young et~al.(2020)Young, Gandhi, Tulsiani, Gupta, Abbeel, and
  Pinto]{young2020visual}
Young, S., Gandhi, D., Tulsiani, S., Gupta, A., Abbeel, P., and Pinto, L.
\newblock Visual imitation made easy.
\newblock \emph{arXiv e-prints}, pp.\  arXiv--2008, 2020.

\bibitem[Zhan et~al.(2020)Zhan, Zhao, Pinto, Abbeel, and
  Laskin]{zhan2020framework}
Zhan, A., Zhao, P., Pinto, L., Abbeel, P., and Laskin, M.
\newblock A framework for efficient robotic manipulation.
\newblock \emph{arXiv preprint arXiv:2012.07975}, 2020.

\bibitem[Zhang et~al.(2017)Zhang, Isola, and Efros]{zhang2017split}
Zhang, R., Isola, P., and Efros, A.~A.
\newblock Split-brain autoencoders: Unsupervised learning by cross-channel
  prediction.
\newblock In \emph{Proceedings of the IEEE Conference on Computer Vision and
  Pattern Recognition}, pp.\  1058--1067, 2017.

\bibitem[Ziebart et~al.(2008)Ziebart, Maas, Bagnell, and
  Dey]{ziebart2008maxent}
Ziebart, B.~D., Maas, A., Bagnell, J.~A., and Dey, A.~K.
\newblock Maximum entropy inverse reinforcement learning.
\newblock In \emph{Proceedings of the 23rd National Conference on Artificial
  Intelligence - Volume 3}, 2008.

\end{thebibliography}
\bibliographystyle{icml2021}

\includeappendixtrue 

\ifincludeappendix

\onecolumn
\appendix
\section*{Appendix}

\section{Extended Background}
\label{section:extended_background}

\paragraph{Soft Actor-Critic} 
The Soft Actor-Critic (SAC)~\citep{haarnoja2018sac} is an off-policy model-free RL algorithm that instantiates an actor-critic framework by learning a state-action value function $Q_\theta$, a stochastic policy $\pi_\theta$  and a temperature $\alpha$ over a discounted infinite-horizon MDP $(\gX, \gA, P, R, \gamma, d_0)$ by optimizing a $\gamma$-discounted maximum-entropy objective~\citep{ziebart2008maxent}. With a slight abuse of notation, we define both the actor and critic learnable parameters by $\theta$. SAC parametrizes  the actor policy $\pi_\theta(\va_t|\vx_t)$ via a $\mathrm{tanh}$-Gaussian defined  as $
\va_t = \mathrm{tanh}(\mu_\theta(\vx_t)+ \sigma_\theta(\vx_t) \epsilon)$, where $\epsilon \sim \gN(0, 1)$, $\mu_\theta$ and $\sigma_\theta$ are parametric mean and standard deviation. The SAC's critic $Q_\theta(\vx_t, \va_t$) is parametrized as an MLP neural network.

The policy evaluation step  learns  the critic $Q_\theta(\vx_t, \va_t)$ network by optimizing the one-step soft Bellman residual:
\begin{align*}
    \gL_Q(\gD) &= \E_{\substack{( \vx_t,\va_t, \vx_{t+1}) \sim \gD \\ \va_{t+1} \sim \pi(\cdot|\vx_{t+1})}}[(Q_\theta(\vx_t, \va_t) - y_t)^2]\text{ and}\\
    y_t &= R(\vx_t, \va_t) + \gamma [Q_{\theta'}(\vx_{t+1}, \va_{t+1}) - \alpha \log \pi_\theta(\va_{t+1}|\vx_{t+1})] ,
\end{align*}
where $\gD$ is a replay buffer of transitions, $\theta'$ is an exponential moving average of $\theta$ as done in~\citep{lillicrap2015ddpg}. SAC uses clipped double-Q learning~\citep{hasselt2015doubledqn,fujimoto2018td3}, which we omit from our notation for simplicity but employ in practice.

The policy improvement step then fits the actor $\pi_\theta(\va_t|\vs_t)$ network by optimizing the following objective:
\begin{align*}
    \gL_\pi(\gD) &= \E_{\vx_t \sim \gD}[ \KL(\pi_\theta(\cdot|\vx_t) || \exp\{\frac{1}{\alpha}Q_\theta(\vx_t, \cdot)\})].
\end{align*}
Finally, the temperature $\alpha$ is learned with the loss:
\begin{align*}
    \gL_\alpha(\gD) &= \E_{\substack{\vx_t \sim \gD \\ \va_t \sim \pi_\theta(\cdot|\vx_t)}}[-\alpha \log \pi_\theta(\va_t|\vx_t) - \alpha \bar\gH],
\end{align*}
where $\bar\gH \in \R$ is the target entropy hyper-parameter that the policy tries to match, which in practice is set to  $\bar\gH=-|\gA|$.
The overall optimization objective of SAC equals to:
\begin{align*}
    \mathcal{L}_\mathrm{SAC}(\gD) &= \gL_\pi(\gD) + \gL_Q(\gD) + \gL_\alpha(\gD).
\end{align*}
We use the $\gL_{\mathrm{SAC}}$ loss as $\gL_{\mathrm{RL}}$ in~\protos.

\section{Experimental Setup}
\label{section:hyperparams}
\subsection{The DeepMind Control Suite Settings}
To benchmark our method we use the DeepMind Control Suite (DMC)~\citep{tassa2018dmcontrol}, a challenging set of image-based continuous control tasks. The episode length of each task  is $1000$ steps, except for \textit{Reach Duplo}, where it is set to $250$. Following~\citet{hafner2019dream}, we set the action repeat hyper-parameter to $2$. An environment observation $\vx \in \gX$ is constructed as a stack of 3 consecutive frames~\citep{mnih2013dqn}, where each frame is an RGB rendering of size $3 \times 84 \times 84$ from the $0^{\text{th}}$ camera, except for the \textit{Quadruped} environment, where we use the $2^{\text{th}}$ camera~\citep{hafner2019dream}, this results into a pixel tensor of size $9\times 84\times 84$. Finally, we divide each pixel's value by $255$ to scale it down to $[0, 1]$ range.

\subsection{Prototypical Representation Learning}
\label{subsection:proto_repr}
{\bf Encoder} We use the convolutional encoder architecture from SAC-AE~\citet{yarats2019improving} to parametrize both the online and target encoders $f_\theta$ and $f_\xi$. This convnet consists of four convolutional layers with $3\times 3$ kernels and $32$ channels. The \texttt{ReLU} activation is applied after each convolutional layer. We use stride to $1$ everywhere, except of the first conv layer, which has stride $2$. The convnet inputs tensors of dimensions $9\times84\times84$ and outputs flatten representations of size $R=32\times35\times35=39200$.\\
{\bf Projector} The online and target projectors $g_\theta$ and $g_\xi$ are just single linear layer $39200 \to 128$ projections.\\
{\bf Predictor} The online $v_\theta$ projector is an MLP with $128 \to 512 \to 128$ architecture and \texttt{ReLU} hidden activations.\\
{\bf Prototypes}~\protos~learns $M=512$ prototypes ($128$-dimensional continuous vectors), where the softmax temperature is set to $\tau=0.1$. To compute the cluster assignments target we employ the Sinkhorn-Knopp algorithm~\citep{cuturi2013sinkhorn}, which performs $n=3$ relaxation iterations per training step.

We train the online network parameters $\theta$ and prototypes $\{\vc_i\}_{i=1}^M$ using stochastic gradient optimization with Adam~\citep{kingma2014adam}, where the learning rate is set to $10^{-4}$ and minibatch size to $512$. The target network parameters $\xi$ being computed as an exponential moving average of $\theta$ with momentum $\tau_{\mathrm{enc}}=0.05$.

\subsection{Entropy-based Intrinsic Reward}
Entropy is being computed in $128$-dimensional latent space that is produced by the online encoder $f_\theta$ and projector $g_\theta$.  We maintain an online candidates queue $Q$ of fixed size $M \times T=512\times 4=2048$, where each of $M=512$ prototypes has exactly $T=4$ candidates. The downstream exploration bonus coefficient is set to $\alpha=0.2$.

\subsection{Soft-Actor Critic Architecture}
\label{subsection:sac_arch}
Our SAC~\citep{haarnoja2018sac} implementation is based on \texttt{github.com/denisyarats/pytorch\_sac}~\cite{yarats2020pytorch_sac} with the following modifications. We add a fully-connected layer of $39200 \to 50$ with \texttt{LayerNorm}~\citep{ba2016layernorm} activation to both actor and critic networks. We also set learning rate to $10^{-4}$, minibatch size to $512$, actor update frequency to $1$, and critic target momentum to $0.01$. 

\subsection{Task-Agnostic Pre-training Setup}
\protos~simultaneously trains representations (see~\cref{subsection:proto_repr}) and exploration RL agent (see~\cref{subsection:sac_arch}) by jointly optimizing $\gL_{\mathrm{SSL}}$ and $\gL_{\mathrm{RL}}$ losses. We perform RL training in the off-policy fashion by maintainng a replay buffer of size $10^5$. The exploration agent first collects $1000$ seed transitions by using a random policy and stores them into the replay buffer. Further training transitions are collected by sampling actions from the exploration policy. One training update to the representations and exploration agent is performed every time a new transition is received. Given the episode's length of $1000$ and fixed action repeat of $2$ we thus perform $500$ training updates per a training episode. In order to learn task-agnostic representations the online encoder $f_\theta$ and prototypes $\{\vc_i\}_{i=1}^M$ are only being updated with the gradients from the $\gL_{\mathrm{SSL}}$ loss, while the gradients from the $\gL_{\mathrm{RL}}$ loss are being blocked.  After pre-training we fix the online encoder $f_\theta$, online projector $g_\theta$ and prototypes $\{\vc_i\}_{i=1}^M$ and prevent them from any further updates during the downstream training.  
\subsection{Task-Specific RL Setup}
During downstream training we train a task RL agent (see~\cref{subsection:sac_arch} for details) on the fixed representations obtained from the encoder $f_\theta$. We also employ the pre-trained prototypes to compute intrinsic reward to combine it together with the true task reward. To ensure initial exploration we initialize the task agent's actor using the exploration actor's weights. 

\subsection{Full List of Hyper-Parameters}
\begin{table}[h]
\caption{\label{table:hyper_params}~\protos~list of hyper-parameters.}
\centering
\begin{tabular}{|l|c|}
\hline
Parameter        & Setting \\
\hline
Replay buffer capacity & $100000$ \\
Seed steps & $1000$ \\
Minibatch size & $512$ \\
Action repeat & $2$ \\
Discount ($\gamma$) & $0.99$ \\
Optimizer & Adam \\
Learning rate & $10^{-4}$ \\
Critic target update frequency & $2$ \\
Critic target EMA momentum ($\tau_{\mathrm{Q}}$) & $0.01$ \\
Actor update frequency & $2$ \\
Actor log stddev bounds & $[-10, 2]$ \\
Encoder target update frequency & $2$ \\
Encoder target EMA momentum ($\tau_{\mathrm{enc}}$) & $0.05$ \\
SAC entropy temperature & $0.1$ \\
Number of prototypes ($M$) & $512$ \\
Number of candidates per prototype ($T$) & $4$ \\
Representation dimensionality ($R$) &  $39200$\\
Latent dimensionality ($D$) & $128$ \\
Softmax temperature ($\tau$) & $0.1$ \\
$k$ in NN & $3$\\
Intrinsic reward coefficient ($\alpha$) & 0.2\\

\hline
\end{tabular}

\end{table}

\subsection{Baselines}
\paragraph{Random} We implement the Random agent baseline based on DrQ~\citep{yarats2021image}. Specifically, during the task-agnostic phase the agent uses a random exploration policy to collect a replay buffer, which is used by DrQ to pre-train the convolutional encoder. During the downstream training, we freeze the encoder convnet and use the downstream task reward to train a DrQ policy on the fixed encoder.
\paragraph{Curiosity} We adapt ICM~\citep{pathak2017curiositydriven} to the off-policy continuous control setting. To facilitate this, we augment DrQ~\citep{yarats2021image} with the ICM module that inputs encoded visual observations and learns forward and inverse dynamics models. The ICM module first projects the visual representations with a linear layer to $50$-dimensional latent vectors. These vectors are then fed into the forward and inverse dynamics, which are parametrized by two layers MLPs with $1024$ hidden units and  \texttt{ReLU} nonlinearities.  As per~\citet{pathak2017curiositydriven}, we use the forward prediction error and an intrinsic signal. We found that normalizing the curiosity reward by a running estimate of its standard deviation and then transforming with the $\mathrm{log\_plus\_one}$ function leads to better performance. During the task-agnostic phase the exploration agent is tasked to optimize the curiosity-based intrinsic reward. After pre-training is completed, we, again, freeze the encoder convnet and use it together with a downstream agent to optimize the target task.

\paragraph{APT} As no original implementation is provided by~\citet{liu2020apt}, we chose to implement APT ourselves follow~\citet{liu2020apt} as close as possible. The only difference to the original implementation is that we freeze the convolution encoder weights of APT during the downstream training to facilitate fair comparison withing our setup. This is in contrast to the setup from~\citet{liu2020apt}, where the encoder fine-tuning is allowed.

\paragraph{Plan2Explore} We obtain results for Plan2Explore from the Table 2 in~\citet{sekar2020planning}. We reemphasize that a direct comparison of our method to Plan2Explore is not meaningful as~\citet{sekar2020planning}~use a different methodology and setup. Specifically,~\citet{sekar2020planning} allows pre-training of the reward model using the task specific rewards during the task-agnostic phase, which leaks the downstream task information. Furthermore, Plan2Explore preserves the replay buffer collected during the task-agnostic phase and uses it during the downstream training, our setup, on the other hand, completely disregards the task-agnostic transitions in the downstream stage. Finally, Plan2Explore allows further fine-tuning of the world-model during the downstream  phase, while we keep  the pre-trained representations fixed. 

\newpage

\section{\protos~Pseudo Code}

\label{section:pseudocode}

\begin{algorithm*}[ht!]

\caption{Pseudocode for \protos~training routine in a PyTorch-like style.}
\label{alg:pseudocode}
\definecolor{codeblue}{rgb}{0.25,0.5,0.5}
\lstset{
  backgroundcolor=\color{white},
  basicstyle=\fontsize{7pt}{7pt}\ttfamily\selectfont,
  columns=fullflexible,
  breaklines=true,
  captionpos=b,
  commentstyle=\fontsize{7pt}{7pt}\color{codeblue},
  keywordstyle=\fontsize{7pt}{7pt},
  escapeinside={(*}{*)}
}
\begin{lstlisting}[language=python]
# C: M prototypes of size D (DxM)
# Q: queue of MxT candidates ((MxT)xD)
# f(*$\color{codeblue} _\theta$*), g(*$\color{codeblue} _\theta$*), v(*$\color{codeblue} _\theta$*): online encoder, projector, and predictor
# f(*$\color{codeblue} _\xi$*), g(*$\color{codeblue} _\xi$*): target encoder and projector
# tau: momentum
# temp: temperature

# sample a minibatch of B transitions without reward from the replay buffer
# (x(*$\color{codeblue} _t$*), a(*$\color{codeblue}_t$*), x(*$\color{codeblue} _{t+1}$*)): state (Bx9x84x84), action (Bx|A|), next state (Bx9x84x84)
for (x(*$_t$*), a(*$_t$*), x(*$_{t+1}$*)) in replay_buffer:
    update_representations(x(*$_t$*), x(*$_{t+1}$*))
    with torch.no_grad():
        r(*$_t$*) = compute_rewards(x(*$_{t+1}$*)) # compute entropy-based task-agnostic reward using the next state x(*$\color{codeblue} _{t+1}$*)
    # decouple representations from RL
    with torch.no_grad():
        y(*$_t$*) = f(*$_\theta$*)(x(*$_t$*)) # obtain representations (BxR)
        y(*$_{t+1}$*) = f(*$_\theta$*)(x(*$_{t+1}$*)) # obtain representations (BxR)
    # train exploration RL agent on an augmented minibatch of B transitions (y(*$\color{codeblue} _t$*), a(*$\color{codeblue}_t$*), r(*$\color{codeblue}_t$*), y(*$\color{codeblue} _{t+1}$*))
    update_rl(y(*$_t$*), a(*$_t$*),  r(*$_t$*),  y(*$_{t+1}$*)) # standard state-based SAC
        
# self-supervised representation learning routine
# x(*$\color{codeblue} _t$*), x(*$\color{codeblue} _{t+1}$*): state (Bx9x84x84) and next state (Bx9x84x84)
def update_representations(x(*$_t$*), x(*$_{t+1}$*)):
    with torch.no_grad():
        C = normalize(C, dim=0, p=2) # normalize prototypes
    # online network
    x(*$_t$*) = aug(x(*$_t$*)) # random-shift view (Bx9x84x84)
    y(*$_t$*) = f(*$_\theta$*)(x(*$_t$*)) # obtain representations (BxR)
    z(*$_t$*) = g(*$_\theta$*)(y(*$_t$*)) # obtain projections (BxD)
    u(*$_t$*) = v(*$_\theta$*)(z(*$_t$*)) # obtain predictions (BxD)
    u(*$_t$*) = normalize(u(*$_t$*), dim=1, p=2) # normalization (BxD)
    p(*$_t$*) = softmax(mm(u(*$_t$*), C) / temp, dim=1) # assignment probabilities (BxM)
    # target network (gradient is blocked)
    with torch.no_grad():
        x(*$_{t+1}$*) = aug(x(*$_{t+1}$*)) # random-shift view (Bx9x84x84)
        y(*$_{t+1}$*) = f(*$_\xi$*)(x(*$_{t+1}$*)) # representation (BxR)
        z(*$_{t+1}$*) = g(*$_\xi$*)(y(*$_{t+1}$*)) # representation (BxD)
        z(*$_{t+1}$*) = normalize(z(*$_{t+1}$*), dim=1, p=2) # normalization (BxD)
        q(*$_{t+1}$*) = sinkhorn(mm(z(*$_{t+1}$*), C) / temp) # target assignments (BxM)
    # cluster assignment loss
    loss = -mean(sum(q(*$_{t+1}$*) * log(p(*$_t$*)), dim=1))
    # SGD update for online network and prototypes
    loss.backward()
    update((*$\theta$*), C)
    # EMA update for the target encoder
    (*$\xi$*) = tau * (*$\xi$*) + (1 - tau) * (*$\theta$*)

# Sinkhorn-Knopp algorithm  
# S: dot products matrix (BxM)
def sinkhorn(S, n=3):
    S = exp(S).T
    S /= sum(S)
    r, c = ones(M) / M, ones(B) / B
    for _ in range(n):
        u = sum(S, dim=1)
        S *= (r / u).unsqueeze(1)
        S *= (c / sum(Q, dim=0)).unsqueeze(0)
    return (S / sum(S, dim=0, keepdim=True)).T # target assignments (BxM)

# entropy-based task-agnostic reward computation
# x: state (Bx9x84x84)
def compute_rewards(x):
    y = f(*$_\theta$*)(x) # obtain representations (BxR)
    z = g(*$_\theta$*)(y) # obtain projections (BxD)
    z = normalize(z, dim=1, p=2) # normalization (BxD)
    w = softmax(mm(z, C).T, dim=1) # candidates softmax probabilities (MxB)
    i = Categorical(w).sample() # one sample per row (Mx1)
    candidates = z[i] # select M candidates (MxD)
    enqueue(Q, candidates) # append the M candidates to the candidates Q, maintain the fixed (MxT) size
    
    # find k-nearest neighbor for each sample in z (BxD) over the candidates queue Q ((MxT)xD)
    dists = norm(z[:, None, :] - Q[None, :, :], dim=-1, p=2) # pairwise L2 distances (Bx(MxT)) between y and Q
    topk_dists, _ = topk(dists, k=3, dim=1, largest=False) # compute topk distances (Bx3)
    r = topk_dists[:, -1:] # rewards (Bx1) are defined as L2 distances to the k-nearest neighbor from Q
    return r
\end{lstlisting}

\end{algorithm*}

\newpage

\section{The PointMass Maze Experiment Details}
\label{section:point_mass}
The U-maze environment is based on the \textit{PointMass Easy} task from DMC~\citep{tassa2018dmcontrol} with the following modifications. First, we add three walls to the MuJoCo model:
\begingroup
    \fontsize{8pt}{8pt}\selectfont
\begin{verbatim}
<default class="wall">
    <geom type="box" material="site"/>
</default>
<geom name="maze_x" class="wall" pos="-.1 0 .02" zaxis="1 0 0"  size=".02 .1 .02"/>
<geom name="maze_neg_x" class="wall" pos=".1 0 .02" zaxis="1 0 0"  size=".02 .1 .02"/>
<geom name="maze_y" class="wall" pos="0 .12 .02" zaxis="0 1 0"  size=".12 .02 .02"/>
\end{verbatim}
\endgroup
We then modify initial state distribution of the point mass from being uniform across the entire $[-0.3, 0.3]\times[-0.3, 0.3]$ grid, to be uniformly distributed across a much smaller region of the state space situated in the top-left corner $[-0.3, -0.15]\times[0.15, 0.3]$. During the task-agnostic stage there is no target location and the agent explores the state space by optimizing the entropy-based intrinsic reward. During the task-specif phase, we place a target location at the center $[0, 0]$ of the grid with radius $0.07$. The agent receives reward of $1$ if it reaches the target location, otherwise it receives no reward. The contrived initial state distribution and sparse reward function make this task extremely hard from the exploration point of view.

\section{The Multitask  Experiment Details}
\label{section:multi_task}
\paragraph{Walker} We add four additional tasks \textit{Run Forward}, \textit{Run Backward}, \textit{Flip Forward}, and \textit{Flip Backward} to the \textit{Walker} environment from DMC that require the agent to run forward/backward, flip forward/backward correspondingly. These tasks are similar to the \textit{Cheetah} tasks from Plan2Explore~\citep{sekar2020planning} that are implemented in \texttt{github.com/ramanans1/dm\_control}.
\paragraph{Reach Duplo} In this set of tasks the agent is required to reach the lego block that is placed in four different fixed locations: \textit{Top Left} $[-0.09, 0.09]$, \textit{Top Right} $[0.09, 0.09]$, \textit{Bottom Left} $[-0.09, -0.09]$, and \textit{Bottom Right} $[0.09, -0.09]$. These task are based on the \textit{Reach Duplo} environment from DMC.
\newpage

\section{Full Results for the Task-Agnostic Pre-training Experiment}
\label{section:full_pretrainig}
We conduct the experiment defined in~\cref{subsection:task_agnostic} on an extended set of $16$ environments from DMC~\citep{tassa2018dmcontrol} and provide them in~\cref{fig:single_task_full}.

\begin{figure*}[h]
    \centering

    \includegraphics[width=\linewidth]{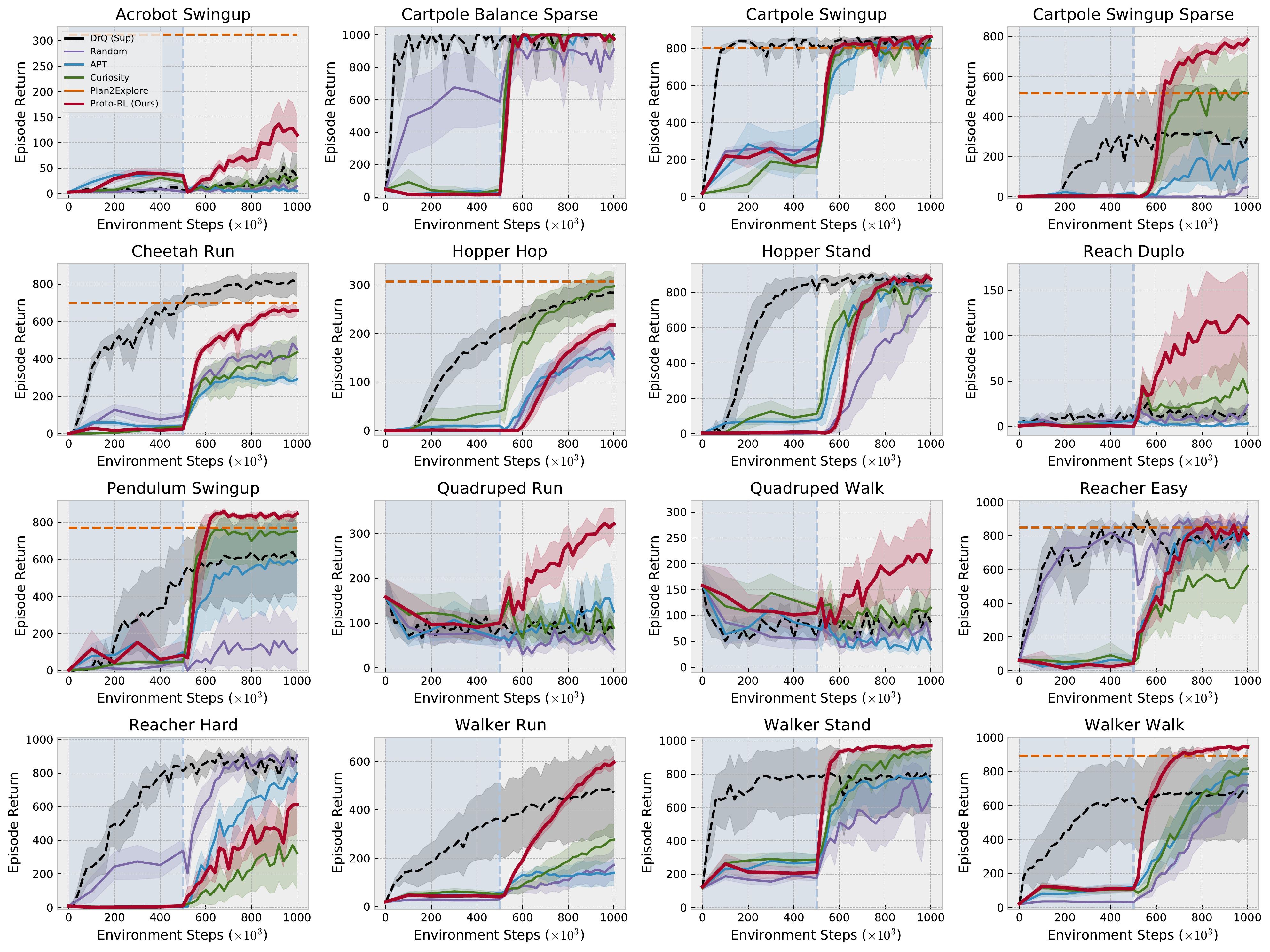}
    \vspace{-7mm}
    \caption{Single task evaluation using a full set of $16$ challenging environments from the DeepMind Control Suite. For each method (except for DrQ and Plan2Explore), we first perform task-agnostic pretraining for 500k environment steps, before introducing task reward and training for a further 500k steps. DrQ uses task reward from the outset. Plan2Explore, being model-based, uses an intermediate methodology.~\protos~consistently beats the baselines and in many cases exceeds the fully supervised approach of DrQ.  }
    \label{fig:single_task_full}
\end{figure*}
\newpage

\section{Full Results for the Efficiency of Task-Agnostic Pre-training Experiment}
\label{section:full_different_pretraining}
In~\cref{fig:different_expl_full} we provide full results of the experiment from~\cref{subsection:eff_of_pretraining}.
\begin{figure}[h]
    \centering

    \includegraphics[width=\linewidth]{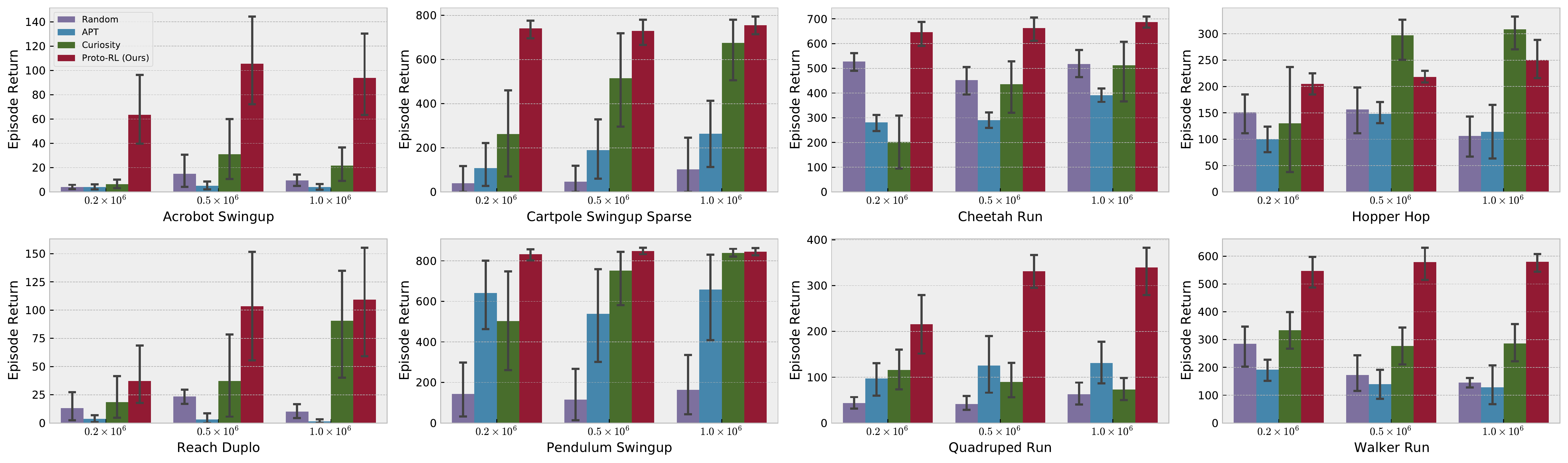}
    \vspace{-5mm}
    \caption{Varying the amount of task-agnostic pre-training on three different tasks from DeepMind Control Suite. Subsequent task-specific training (on top of the frozen representation) uses 500k steps. \protos~is able to explore state space sufficiently within 200k steps to learn representations that can support downstream tasks.}
    \label{fig:different_expl_full}
\end{figure}

\section{Full Results for the Downstream Exploration Experiment}
\label{section:full_downstream_expl}
In~\cref{fig:downstream_expl_full} we provide full results of the experiment from~\cref{subsection:downstream_exploration}.

\begin{figure}[h]
    \centering

    \includegraphics[width=\linewidth]{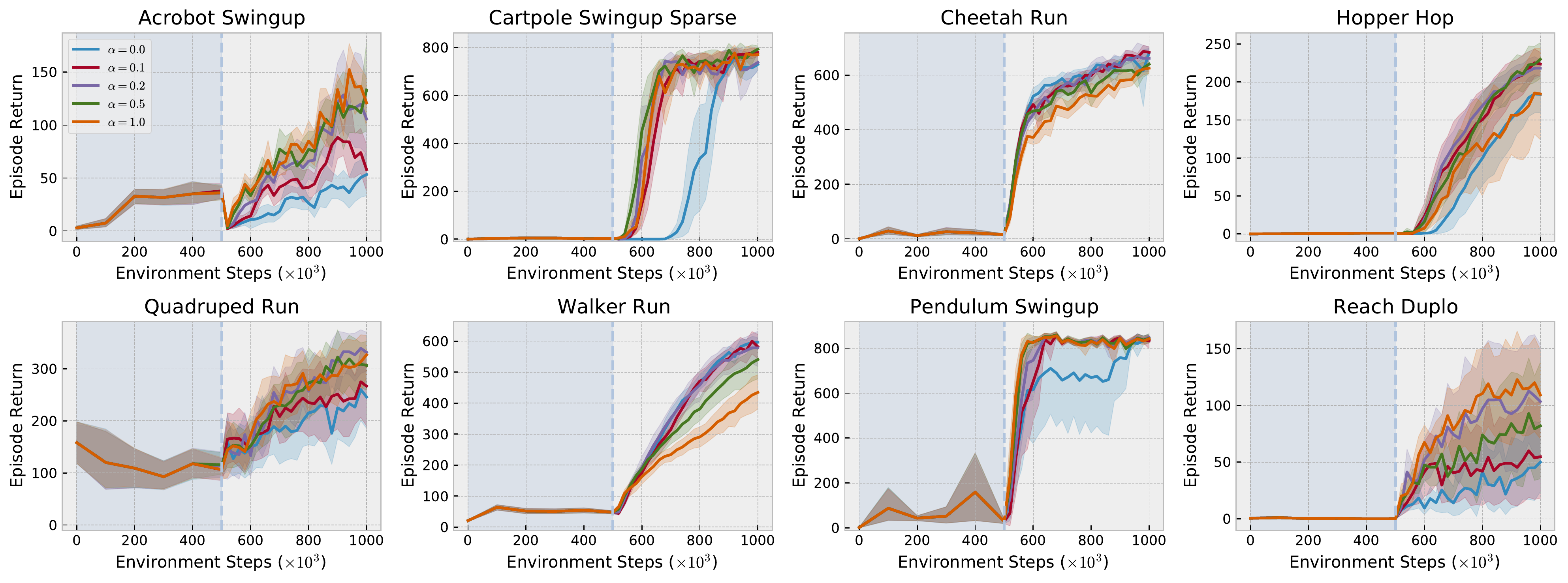}
    \vspace{-5mm}
    \caption{Evaluation regime from~\cref{fig:single_task} for \protos~but varying the balance between exploration and downstream reward using hyper-parameter $\alpha$. We see that $\alpha > 0$ facilitates learning, especially in the sparse reward tasks such as \textit{Cartpole Swingup Sparse} and \textit{Reach Duplo}.}
    \label{fig:downstream_expl_full}
\end{figure}

\section{Full Results on the Downstream Performance without Pre-training Experiment}
\label{section:no_pretraining}
In this section, we demonstrates that Proto-RL's superior performance does not exclusively come from faster downstream training, but actually significantly depends on our novel task-agnostic pre-training scheme.~\cref{fig:no_pretraining} proves that Proto-RL without prior task-agnostic pre-training is incapable of matching the performance of the fully fledged Proto-RL.

\begin{figure}[t!]
    \centering

    \includegraphics[width=\linewidth]{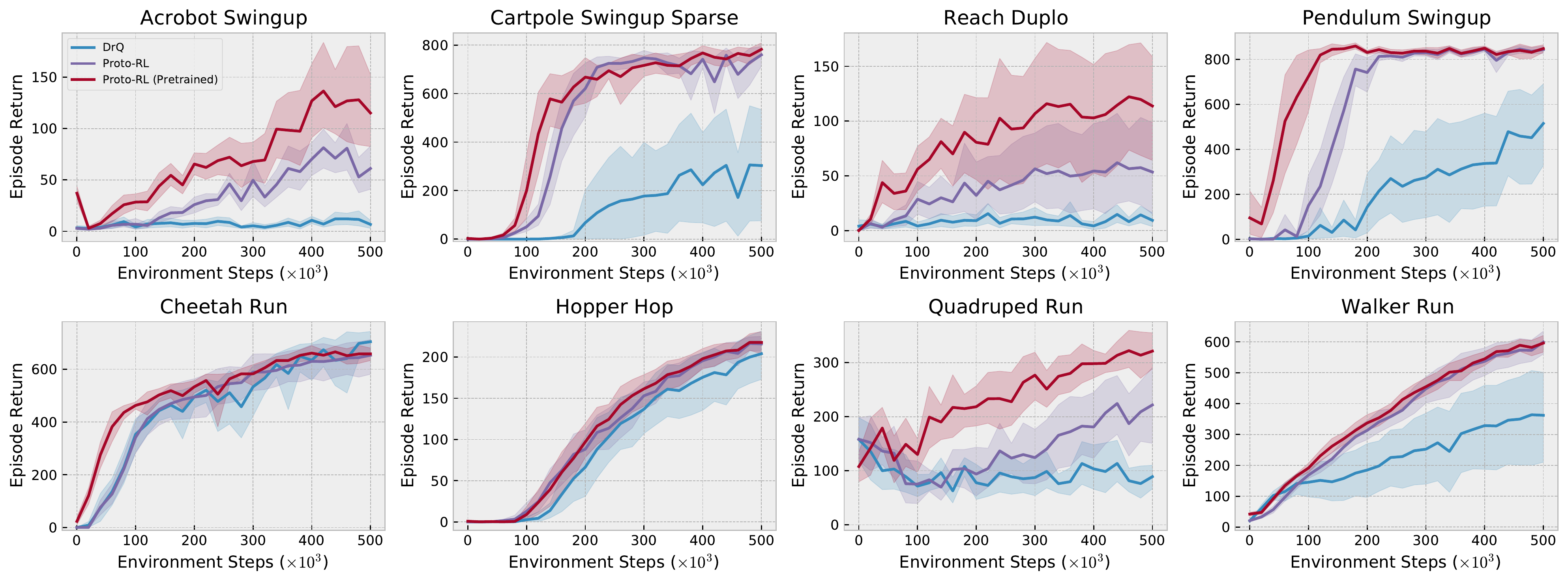}
    \vspace{-5mm}
    \caption{Single task evaluation on a set of $8$ challenging environments from the DeepMind Control Suite without prior task-agnostic pre-training. In this setting, both Proto-RL and DrQ~\cite{yarats2021image}  optimize the downstream objective right from the first step of training. While Proto-RL outperforms DrQ, it still demonstrates inferior results compared to Proto-RL (Pretrained), a variant that foregoes a stage of prior task-agnostic pre-training for 500k environment steps. This result showcases Proto-RL's ability to learn useful representations without supervision.}
    \label{fig:no_pretraining}
\end{figure}
\fi

\end{document}